\documentclass[sigconf, screen, nonacm]{acmart}
\AtBeginDocument{%
  }

\renewcommand\footnotetextcopyrightpermission[1]{}

\theoremstyle{definition}

\newtheorem{proposition}{Proposition}
\newtheorem{theorem}{Theorem}

\usepackage{wrapfig}

\usepackage{booktabs}
\usepackage{multirow}
\usepackage{graphicx}
\usepackage{colortbl}

\definecolor{RankFirst}{HTML}{D5E8D4}   
\definecolor{RankSecond}{HTML}{FFF2CC}  
\definecolor{RankThird}{HTML}{DAE8FC}   
\usepackage{pifont}
\newcommand{\cmark}{\textcolor{green}{\ding{51}}}
\newcommand{\xmark}{\textcolor{red}{\ding{55}}}
\makeatletter
\AtBeginDocument{%
  \fancypagestyle{standardpagestyle}{%
    \fancyhf{}
    \renewcommand{\headrulewidth}{\z@}
    \renewcommand{\footrulewidth}{\z@}
    \fancyfoot[C]{\if@ACM@printfolios\footnotesize\thepage\fi}}}
\makeatother

\begin{document}

\title{TransSplat: Unbalanced Semantic Transport for Language-Driven 3DGS Editing}

\author{Yanhui Chen}
\affiliation{%
  \institution{Guangdong University of Technology}
  \city{Guangzhou}
  \country{China}}
\email{chenyanhui91@mails.gdut.edu.cn}

\author{Jiahong Li}
\affiliation{%
  \institution{Guangdong University of Technology}
  \city{Guangzhou}
  \country{China}}
\email{488241817@mails.gdut.edu.cn}

\author{Jingchao Wang}
\affiliation{%
  \institution{Peking University}
  \city{Beijing}
  \country{China}}
\email{ethanwangjc@163.com}

\author{Junyi Lin}
\affiliation{%
  \institution{Guangdong University of Technology}
  \city{Guangzhou}
  \country{China}}
\email{3123001378@mail2.gdut.edu.cn}

\author{Zixin Zeng}
\affiliation{%
  \institution{Guangdong University of Technology}
  \city{Guangzhou}
  \country{China}}
\email{zengzixin@mails.gdut.edu.cn}

\author{Yang Shi}
\affiliation{%
  \institution{Guangdong University of Technology}
  \city{Guangzhou}
  \country{China}}
\email{sudo.shiyang@gmail.com}

\renewcommand{\shortauthors}{Chen et al.}

\begin{abstract}
Language-driven 3D Gaussian Splatting (3DGS) editing provides a more convenient approach for modifying complex scenes in VR/AR. Standard pipelines typically adopt a two-stage strategy: first editing multiple 2D views, and then optimizing the 3D representation to match these edited observations. Existing methods mainly improve view consistency through multi-view feature fusion, attention filtering, or iterative recalibration. However, they fail to explicitly address a more fundamental issue: the semantic correspondence between edited 2D evidence and 3D Gaussians. To tackle this problem, we propose TransSplat, which formulates language-driven 3DGS editing as a multi-view unbalanced semantic transport problem. Specifically, our method establishes correspondences between visible Gaussians and view-specific editing prototypes, thereby explicitly characterizing the semantic relationship between edited 2D evidence and 3D Gaussians. It further recovers a cross-view shared canonical 3D edit field to guide unified 3D appearance updates. In addition, we use transport residuals to suppress erroneous edits in non-target regions, mitigating edit leakage and improving local control precision. Qualitative and quantitative results show that, compared with existing 3D editing methods centered on enhancing view consistency, TransSplat achieves superior performance in local editing accuracy and structural consistency.

\end{abstract}



\begin{teaserfigure}
\centering
\includegraphics[width=0.96\textwidth]{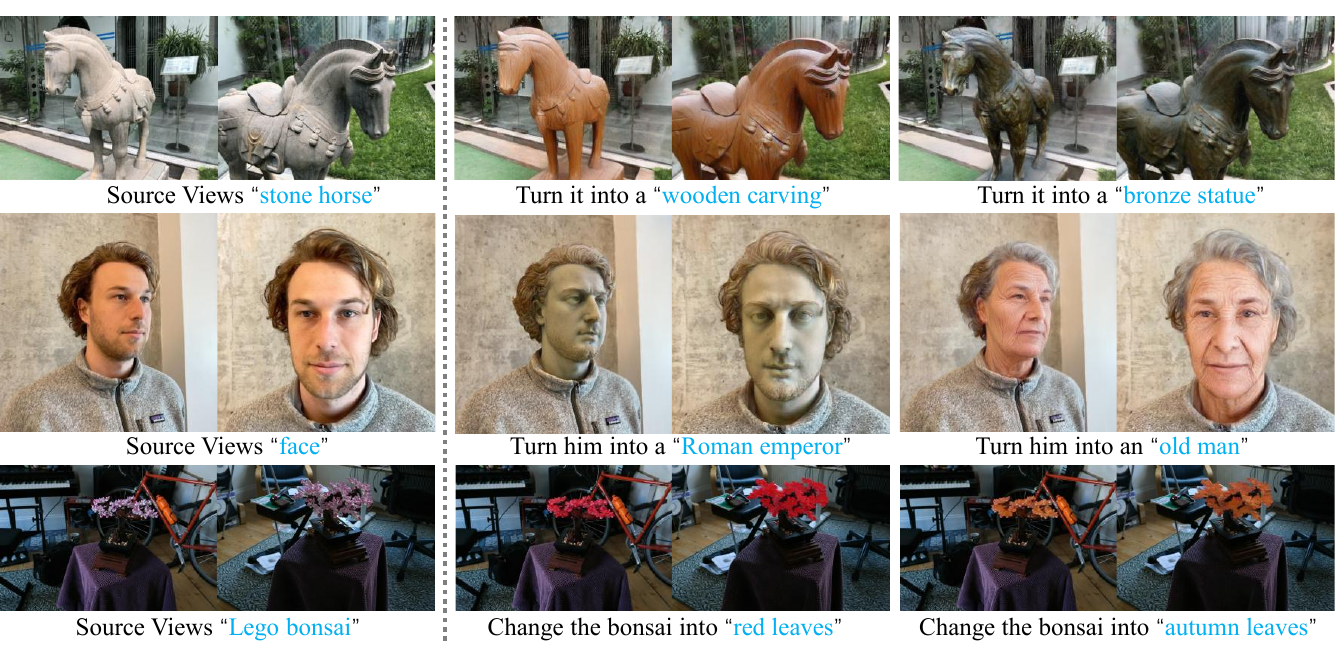}
\caption{Editing results of TransSplat. Relying only on text instructions, our method enables realistic, precise, and view-consistent editing of pretrained 3D Gaussian Splatting scenes.}
\Description{Representative editing results of TransSplat on pretrained 3D Gaussian Splatting scenes under text-guided instructions.}
\label{fig:teaser}
\end{teaserfigure}

\maketitle

\section{Introduction}

Text-driven 3D scene editing aims to directly modify 3D content through natural language instructions and has become an important research direction in recent years. Recent benchmark studies further show that vision-language models can still exhibit nontrivial process-level reasoning errors under multimodal inputs \cite{shi2026mmerror}, which highlights the importance of establishing reliable language-grounded correspondences during editing. Instruct-NeRF2NeRF (IN2N) \cite{haque2023instruct} was the first to systematically introduce the 2D-to-3D pipeline, which first edits rendered views with 2D diffusion models and then propagates the edited results back to optimize the underlying 3D representation \cite{haque2023instruct, brooks2023instructpix2pix, rombach2022high}. Inspired by this paradigm, a large body of subsequent work \cite{wang2022clip, poole2022dreamfusion, zhang2024text2nerf, zhuang2023dreameditor} extended it to different types of neural 3D representations. As the advantages of 3D Gaussian Splatting (3DGS) in rendering efficiency and local update capability have become increasingly evident, an increasing number of methods \cite{kerbl20233d, chen2024gaussianeditor, wang2024gaussianeditor} have migrated this paradigm to 3DGS scene editing. Since such methods fundamentally rely on multi-view 2D editing results to constrain a shared 3D optimization process, maintaining cross-view consistency has gradually become one of the central challenges in this area.

To address this issue, existing studies \cite{wu2024gaussctrl, dong2023vica, khalid20243dego} have mainly sought to improve the consistency of multi-view editing results from different perspectives. For example, some methods enhance appearance coherence across views through key-view optimization, attention aggregation, or cross-view feature fusion, while others integrate multi-view editing information into a unified 3D representation through strategies such as depth constraints, geometric guidance, or iterative recalibration. These methods have achieved promising results on tasks such as style transfer, color replacement, and relatively stable appearance modification, showing that the proper use of multi-view 2D editing results can indeed improve the quality and practicality of text-driven 3D editing.

\begin{figure}[t]
\centering
\includegraphics[width=1\linewidth]{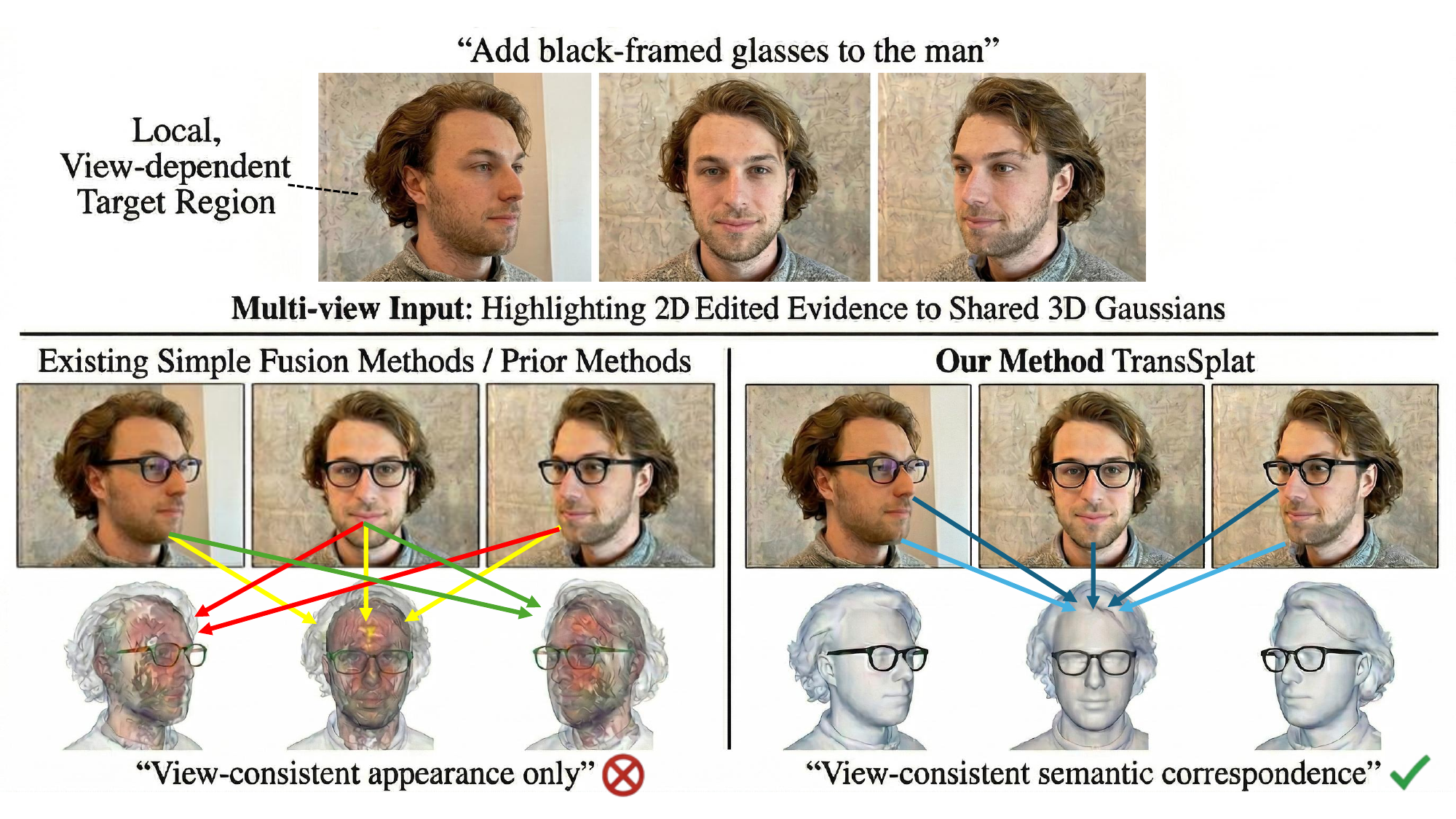}
\caption{Simple multi-view fusion can produce edited results with similar appearance across views, yet it may still assign inconsistent editing semantics to the same 3D Gaussians. Our method instead estimates view-aware correspondences between edited 2D evidence and shared 3D Gaussians, leading to more consistent semantic updates across viewpoints.}
\Description{A three-part motivation figure comparing input views, simple fusion with inconsistent 3D semantic assignments, and the proposed method with consistent semantic correspondence.}
\label{fig_motivation}
\end{figure}

However, although existing methods alleviate cross-view appearance inconsistency to some extent, they still fail to explicitly address a more fundamental challenge: the semantic correspondence between edited 2D evidence and the shared 3D Gaussians. When a text instruction involves local material changes, structural modifications, or object identity transformations, different views usually provide only partial and noisy evidence. As a result, a method may still produce visually similar results across views while assigning inconsistent editing semantics to the same 3D region. This issue becomes particularly severe in scenarios involving non-rigid changes, partial occlusion, and fine-grained local control. As illustrated in Fig.~\ref{fig_motivation}, the mapping between 2D edited regions and 3D Gaussians is inherently many-to-many, incomplete, and non-mass-preserving, which also explains why existing methods based only on fusion, filtering, or recalibration cannot fundamentally resolve this problem.

Based on these observations, we propose TransSplat, which formulates language-driven 3DGS editing as a multi-view unbalanced semantic transport problem \cite{peyre2019computational, chizat2018unbalanced, chizat2016scaling, cuturi2014fast}. Unlike existing methods that mainly improve multi-view appearance consistency through post-edit fusion, filtering, or recalibration, TransSplat focuses on a more fundamental question: how multi-view 2D editing evidence can induce consistent, stable, and reasonable semantic assignments over a shared 3D Gaussian representation. To this end, we introduce a compact semantic latent for each Gaussian to encode not only appearance attributes such as color and opacity, but also editing semantics shared across views. We further extract text-relevant editing prototypes from the 2D features of each edited view to summarize the most critical local evidence. Based on these representations, we establish transport relations between visible Gaussians and view-specific prototypes, explicitly modeling the connection between edited 2D evidence and 3D Gaussians through projection geometry, semantic discrepancy, and appearance discrepancy. Since single-view editing signals are often incomplete and noisy, we do not directly rely on local results. Instead, we integrate semantic information from multiple views into a cross-view shared canonical 3D edit field, which serves as a unified 3D semantic representation of the target edit and guides consistent 3D appearance updates. In addition, we use transport residuals to constrain erroneous changes in non-target regions, thereby mitigating edit leakage and improving the precision and stability of local editing. Overall, TransSplat remains lightweight and compatible with existing 2D-to-3D editing pipelines, while providing a more interpretable framework for language-driven 3DGS editing that is also more amenable to theoretical analysis.

The contributions of this work are listed below.

(1) We formulate language-driven 3D Gaussian editing as a multi-view semantic correspondence problem and propose a transport based solution that explicitly links edited 2D evidence to shared 3D Gaussians.

(2) We introduce a canonical 3D edit field obtained from confidence-weighted barycentric fusion of view-wise transported semantics together with transport guided leakage suppression for stable local updates.

(3) We provide a concise theory that establishes uniqueness of the view-wise unbalanced transport solution and stability of the canonical edit field under noisy per view observations.

(4) Under the same benchmark setting as prior 3D editing work, we achieve better quantitative metrics and qualitative results on editing across eight scenes.


\section{Related Work}

\noindent\textbf{2D Image Editing.}
In recent years, diffusion-based 2D image editing methods have achieved remarkable progress in tasks such as text control, local modification, subject customization, and conditional guidance \cite{brooks2023instructpix2pix, hertz2022prompt, zhang2023adding, mokady2023null}. Among them, text-to-image (T2I) diffusion models and their derived editing methods have gradually become the dominant paradigm in this area, providing a unified foundation for high-quality and controllable image generation and editing \cite{rombach2022high, radford2021learning, lu2026chordedit, kulikov2025flowedit}. In the direction of instruction-driven editing, InstructPix2Pix enables direct modification of input images according to natural language instructions by fine-tuning Stable Diffusion on a large-scale instruction-based editing dataset. The development of these 2D editing methods has provided a critical foundation for subsequent text-driven 3D editing: on the one hand, they have significantly improved the realism and controllability of editing results; on the other hand, they have supplied directly usable image editors and strong generative priors for the 2D-to-3D editing paradigm of first editing 2D views and then propagating the results back to optimize the 3D representation.

\noindent\textbf{3D Editing in NeRF.}
NeRF served as an important early representation for text-driven 3D editing. Related methods mainly introduce the priors of 2D generative models into the 3D optimization process through score distillation, text-guided optimization, or sparse-view supervision \cite{mildenhall2021nerf, poole2022dreamfusion, wang2022clip, zhang2024text2nerf}. Among them, Instruct-NeRF2NeRF (IN2N) was the first to systematically establish a 2D-to-3D editing paradigm: it first renders multi-view images, then edits them with a 2D image editor, and finally uses the edited results to update the underlying 3D representation. Most subsequent works followed this render--edit--optimize pipeline and extended it to different types of neural 3D representations.

\noindent\textbf{3D Editing in 3DGS.}
As the advantages of 3D Gaussian Splatting (3DGS) in rendering efficiency, representational capability, and local updates have become increasingly evident, more and more works have begun to migrate text-driven 3D editing from NeRF to 3DGS scene representations \cite{kerbl20233d}. Existing 3DGS editing methods generally aim to improve multi-view consistency and the precision of local modifications, with their core ideas mainly reflected in region-aware updates, depth or geometric constraints, multi-view editing evidence fusion, and attention-guided optimization \cite{kerbl20233d, chen2024gaussianeditor, wang2024gaussianeditor, wu2024gaussctrl, chen2024dge, lee2025editsplat}. These methods have significantly improved editing quality in tasks such as material changes, style transfer, and object-level modification, and have advanced the efficiency and controllability of text-driven 3DGS editing.

\begin{figure*}[t]
    \centering
    \includegraphics[width=0.98\linewidth]{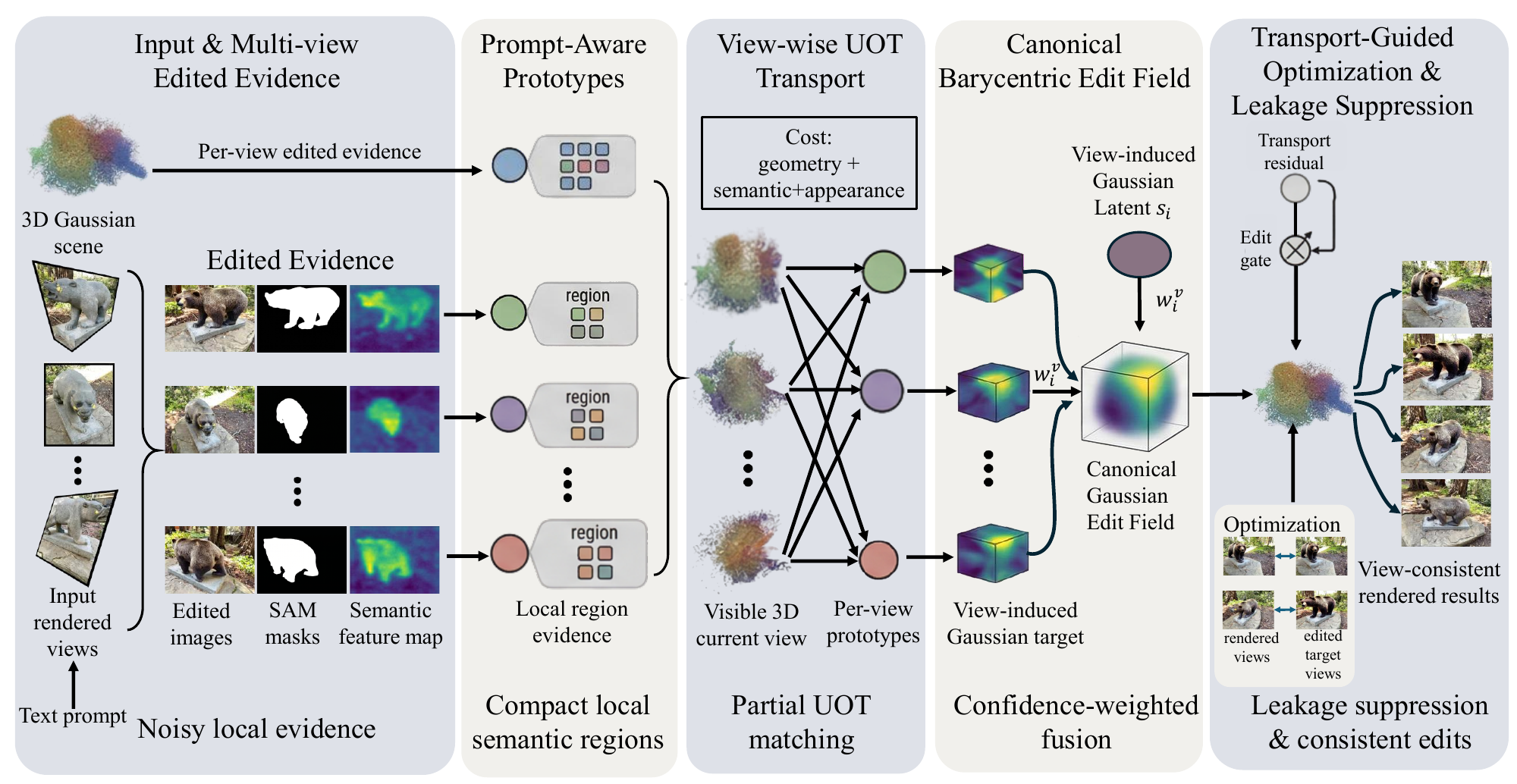}
    \caption{\textbf{Overview of TransSplat.} We first extract editing evidence from multi-view edited results and compress it into a set of prompt-aware local prototypes. We then establish semantic correspondences between visible 3D Gaussians and per-view prototypes through view-wise unbalanced optimal transport, yielding view-induced Gaussian editing targets. These targets are further fused via confidence-weighted barycentric aggregation to construct a unified canonical Gaussian edit field. Finally, transport-guided optimization updates the 3D Gaussian appearance while suppressing edit leakage in weakly supported regions, producing view-consistent edited results.}
    \Description{Overview of the TransSplat pipeline for language-driven 3D Gaussian editing.}
    \label{fig:overview}
\end{figure*}

\section{Method}
\subsection{Problem Setup}

We consider a scene represented by a set of $N$ Gaussians, denoted by $\mathcal{G}=\{g_i\}_{i=1}^{N}$, where each Gaussian is written as $g_i=(x_i,\Sigma_i,c_i,\alpha_i,s_i)$. In this representation, $x_i\in\mathbb{R}^{3}$ denotes the Gaussian center, $\Sigma_i\in\mathbb{S}_{++}^{3}$ denotes the covariance matrix, $c_i\in[0,1]^{3}$ and $\alpha_i\in(0,1)$ denote the color and opacity, respectively, and $s_i\in\mathbb{R}^{d}$ is a learnable semantic latent of dimension $d$. The scene is observed from $V$ input views $\{(I_v,\Pi_v)\}_{v=1}^{V}$, where $I_v$ is the input image and $\Pi_v$ is the camera projection. Given a text prompt $y$, a 2D editor produces, for each optimization view, an edited target image $\hat I_v$, an edit attention map $A_v$, and an edited feature map $E_v$. We denote by $\mathcal{I}_v\subseteq\{1,\ldots,N\}$ the index set of Gaussians visible in view $v$, and use $R(\mathcal{G},\Pi_v)$ to denote the renderer.

Our goal is to edit $\mathcal{G}$ such that the rendered image $R(\mathcal{G},\Pi_v)$ matches the desired semantics in $\hat I_v$ while preserving cross-view consistency. The main difficulty is that each edited view provides only partial evidence for the final 3D edit, and this evidence is often noisy and uneven across views. As a result, directly propagating per-view 2D editing results back to the 3D representation is generally insufficient. Instead, we seek a formulation that first establishes a stable view-wise correspondence operator and then integrates the resulting multi-view evidence into a canonical 3D edit field.

\subsection{Prompt-Aware Prototype Extraction}

Directly establishing correspondences between each edited view and the visible Gaussians at the pixel level would make the computation grow rapidly with image resolution, while pixel-level evidence itself contains abundant fine-grained noise that can easily destabilize the subsequent semantic assignment process. For language-driven 3D editing, what truly matters is not every small pixel fluctuation, but the local semantic regions that capture where the current instruction intends to edit and what semantic change it intends to impose in the current view. Motivated by this observation, we do not directly use dense pixel-level editing evidence. Instead, we compress each edited view into a compact set of prompt-aware prototypes, which summarize the most important local editing signals in that view \cite{kirillov2023segment, oquab2023dinov2, liu2024grounding}.

Let $E_v(p)\in\mathbb{R}^{d}$ denote the edited semantic feature at pixel $p$ in view $v$. We first perform confidence-aware clustering over the effective support region of the attention map $A_v$ and partition the current view into $M_v$ local regions $\{\Omega_m^v\}_{m=1}^{M_v}$. Each local region is summarized as a prototype triplet
\begin{equation}
p_m^v=(u_m^v,e_m^v,b_m^v),
\end{equation}
where $u_m^v$ denotes the representative spatial location of that region, $e_m^v$ denotes its aggregated semantic feature, and $b_m^v$ denotes the target mass carried by that region. To obtain these quantities, we first compute the attention-weighted statistics over each local region,
\begin{equation}
\kappa_m^v=\sum_{p\in\Omega_m^v} A_v(p), \hspace{0.6em}
\nu_m^{v,u}=\sum_{p\in\Omega_m^v} A_v(p)\,p, \hspace{0.6em}
\nu_m^{v,E}=\sum_{p\in\Omega_m^v} A_v(p)\,E_v(p)
\end{equation}
\begin{equation}
u_m^v=\frac{\nu_m^{v,u}}{\kappa_m^v}, \qquad
e_m^v=\frac{\nu_m^{v,E}}{\kappa_m^v}, \qquad
b_m^v=\kappa_m^v.
\end{equation}
After this compression step, each prototype no longer corresponds to an individual pixel, but instead represents a local edited region that is relatively concentrated in space and coherent in semantics. The original editing evidence is therefore reorganized into a view-specific target measure that is more compact, more stable, and more suitable for subsequent optimization.

Relying only on spatial location and semantic averages is still not sufficiently robust in regions with complex local structure or pronounced texture variation. We therefore further equip each prototype with a local appearance descriptor. Let $\Phi_v(p)\in\mathbb{R}^{d_a}$ denote the pixel-level appearance feature. The appearance statistics of each local region $\Omega_m^v$ are obtained through attention-weighted aggregation and are then used to construct the appearance descriptor associated with the prototype. This appearance cue is not introduced to dominate the transport itself, but to provide an additional local visual reference beyond geometry and semantics, thereby reducing mismatches between regions that are close in image space but not truly aligned in visual content. After this stage, the editing evidence in each view is transformed from a dense and noisy pixel collection into a compact representation composed of a small number of high-confidence local prototypes, which provides a much more stable input for the subsequent view-wise semantic transport stage.

\subsection{View-Wise Unbalanced Semantic Transport}
After obtaining the prompt-aware prototypes for each view, the next question is how the local editing evidence in the current view should be assigned to the visible 3D Gaussians. To address this, we treat the visible Gaussians in view $v$ as a source measure and the extracted prototype set in that view as a target measure, and establish a view-wise unbalanced semantic transport process between them. For each visible Gaussian $g_i$ in view $v$, we define a source mass $a_i^v$ according to its visibility weight and opacity, which characterizes its capacity to carry the editing signal under the current view. At the same time, we define a projected appearance descriptor $\phi_i^v$ for each visible Gaussian and construct a corresponding appearance descriptor $\psi_m^v$ for each prototype using attention-weighted aggregation of local appearance features. The resulting transport relation no longer depends only on 2D position, but jointly considers geometric locality, semantic consistency, and local appearance similarity, allowing editing evidence to be mapped more reliably from the 2D view to the shared 3D Gaussian representation.

Under this formulation, the transport cost from Gaussian $g_i$ to prototype $p_m^v$ is determined by geometric, semantic, and appearance terms,
\begin{equation}
C_{im}^v = \lambda_{\mathrm{geo}} \lVert \Pi_v(x_i) - u_m^v \rVert_2^2
+\lambda_{\mathrm{sem}} \Bigl(1 - \frac{\langle s_i, e_m^v \rangle}{\lVert s_i \rVert_2 \, \lVert e_m^v \rVert_2 + \delta} \Bigr)
+\lambda_{\mathrm{app}} \lVert \phi_i^v - \psi_m^v \rVert_2^2,
\end{equation}
where the geometric term encourages locality in the image plane, the semantic term measures the agreement between the Gaussian semantic latent and the prototype semantic feature, and the appearance term further reduces mismatches between regions that are spatially close but not truly aligned in visual content. Based on this cost, we solve an entropically regularized unbalanced transport problem within the current view,
\begin{equation}
T_v^{\star}
=
\arg\min_{T \ge 0}
\Bigl[
\langle C_v, T \rangle
+ \varepsilon \, \mathrm{KL}(T \| a_v b_v^{\top})
+ \tau_s \, \mathrm{KL}(T \mathbf{1} \| a_v)
+ \tau_t \, \mathrm{KL}(T^{\top} \mathbf{1} \| b_v)
\Bigr].
\end{equation}
The vectors $a_v$ and $b_v$ denote the source-mass vector and the prototype-mass vector, respectively, while $\varepsilon$, $\tau_s$, and $\tau_t$ control the strength of entropy regularization and marginal relaxation. The use of an unbalanced formulation, rather than a standard balanced one, is motivated by the nature of language-driven editing itself: an instruction may introduce new local semantics, weaken existing ones, or even remove certain structures and textures altogether. Allowing a mismatch between source and target mass therefore better reflects the actual editing process.

Once the optimal transport plan $T_v^\star$ is obtained, each visible Gaussian receives a view-induced local semantic editing target together with a transport confidence that reflects how strongly the current view supports editing that Gaussian,
\begin{equation}
y_i^v = \frac{\sum_{m=1}^{M_v} T_{im}^{v\star} e_m^v}{\sum_{m=1}^{M_v} T_{im}^{v\star} + \delta},
\qquad
w_i^v = \sum_{m=1}^{M_v} T_{im}^{v\star}.
\end{equation}
The former specifies the semantic direction toward which Gaussian $g_i$ should be updated under view $v$, while the latter measures the strength of support provided by that view. Through this step, the originally sparse and noisy editing evidence in the 2D image space is converted into structured semantic targets defined over the visible Gaussians, which in turn provides a unified intermediate representation for the subsequent cross-view fusion and recovery of a canonical edit field.

\subsection{Canonical Barycentric Edit Field}

A view-local semantic target alone is insufficient for the final 3D edit, because the same Gaussian is typically visible in multiple views, while the edited scene must ultimately maintain a single consistent semantic state. In other words, the quantity $y_i^v$ obtained in the previous stage only reflects what view $v$ suggests Gaussian $g_i$ should become, and these view-specific targets may vary in reliability or even conflict with one another. We therefore require a unified fusion mechanism that aggregates such local semantic evidence into a cross-view shared 3D editing target.

For each Gaussian index $i$, we denote by $\mathcal{V}(i)$ the set of views in which $g_i$ is visible and satisfies $w_i^v > 0$. Based on the transport confidence from each view, we define the normalized weight
\begin{equation}
\omega_i^v = \frac{w_i^v}{\sum_{v' \in \mathcal{V}(i)} w_i^{v'} + \delta}.
\end{equation}
This weight reflects the relative support that different views provide for editing Gaussian $g_i$: a view with clearer and more reliable editing evidence receives a larger weight and therefore contributes more to the subsequent fusion. Rather than directly averaging the view-wise semantic targets, we recover the canonical semantic target of Gaussian $g_i$ by solving the following regularized barycentric problem:
\begin{equation}
z_i^{\star} = \arg\min_{z \in \mathbb{R}^{d}} \sum_{v \in \mathcal{V}(i)} \omega_i^v \lVert z - y_i^v \rVert_2^2 + \rho \lVert z - s_i \rVert_2^2.
\end{equation}

The first term encourages the canonical target $z_i^\star$ to remain consistent with the view-induced semantic targets $\{y_i^v\}$ while weighting their contributions according to confidence. The second term regularizes the solution toward the current Gaussian semantic latent $s_i$, preventing excessive drift when the multi-view evidence is sparse or unstable. The regularization strength is controlled by $\rho \ge 0$. The resulting $z_i^\star$ defines the canonical 3D edit field for Gaussian $g_i$, namely a cross-view shared semantic editing target in 3D. Compared with direct averaging, this barycentric formulation makes the fusion objective explicit, handles unequal support from different views more naturally, and provides a clean mathematical basis for the stability analysis presented later.

\begin{table*}[t]
\centering
\caption{\textbf{Main quantitative comparison across 8 scenes.} The Avg row reports the mean over all scenes for each metric. The best, second, and third results are highlighted with \colorbox{RankFirst}{green}, \colorbox{RankSecond}{yellow}, and \colorbox{RankThird}{blue} backgrounds.}
\label{tab:main_results}
\resizebox{0.98\textwidth}{!}{%
\begin{tabular}{@{}l | cc | cc | cc | cc@{}}
\toprule
\multirow{2}{*}{\textbf{Scenes}} & \multicolumn{2}{c|}{\textbf{Gaussctrl}} & \multicolumn{2}{c|}{\textbf{DGE}} & \multicolumn{2}{c|}{\textbf{EditSplat}} & \multicolumn{2}{c}{\textbf{Ours (TransSplat)}} \\
\cmidrule(lr){2-3} \cmidrule(lr){4-5} \cmidrule(lr){6-7} \cmidrule(lr){8-9}
& \textbf{CLIP-sim}$\uparrow$ & \textbf{CLIP-dir}$\uparrow$ & \textbf{CLIP-sim}$\uparrow$ & \textbf{CLIP-dir}$\uparrow$ & \textbf{CLIP-sim}$\uparrow$ & \textbf{CLIP-dir}$\uparrow$ & \textbf{CLIP-sim}$\uparrow$ & \textbf{CLIP-dir}$\uparrow$ \\
\midrule
\rowcolor{gray!15} \multicolumn{9}{c}{\textbf{Part I: Per-Scene Evaluation}} \\
\midrule
Bear & \cellcolor{RankThird}0.2583 & -0.0051 & \cellcolor{RankSecond}0.2872 & \cellcolor{RankSecond}0.1814 & 0.2266 & \cellcolor{RankThird}0.1539 & \cellcolor{RankFirst}\textbf{0.3040} & \cellcolor{RankFirst}\textbf{0.2520} \\
Bicycle & \cellcolor{RankThird}0.1847 & \cellcolor{RankThird}0.0354 & 0.1555 & -0.0035 & \cellcolor{RankSecond}0.2057 & \cellcolor{RankSecond}0.1013 & \cellcolor{RankFirst}\textbf{0.2469} & \cellcolor{RankFirst}\textbf{0.1490} \\
Bonsai & 0.2081 & -0.0258 & \cellcolor{RankThird}0.2101 & \cellcolor{RankThird}0.0012 & \cellcolor{RankSecond}0.2164 & \cellcolor{RankSecond}0.0255 & \cellcolor{RankFirst}\textbf{0.2469} & \cellcolor{RankFirst}\textbf{0.0669} \\
Face & 0.2188 & \cellcolor{RankThird}0.1051 & \cellcolor{RankThird}0.2345 & 0.0676 & \cellcolor{RankSecond}0.2464 & \cellcolor{RankSecond}0.1133 & \cellcolor{RankFirst}\textbf{0.2539} & \cellcolor{RankFirst}\textbf{0.1365} \\
Fangzhou & \cellcolor{RankThird}0.2353 & \cellcolor{RankFirst}0.2069 & 0.1351 & -0.0552 & \cellcolor{RankSecond}0.2445 & \cellcolor{RankThird}0.1763 & \cellcolor{RankFirst}\textbf{0.2461} & \cellcolor{RankSecond}\textbf{0.1913} \\
Garden & \cellcolor{RankFirst}0.2493 & \cellcolor{RankThird}0.1291 & 0.2251 & 0.0320 & \cellcolor{RankThird}0.2294 & \cellcolor{RankSecond}0.1363 & \cellcolor{RankSecond}\textbf{0.2441} & \cellcolor{RankFirst}\textbf{0.1594} \\
Person & 0.2436 & \cellcolor{RankThird}0.1941 & \cellcolor{RankThird}0.2610 & \cellcolor{RankFirst}0.2313 & \cellcolor{RankFirst}0.2652 & \cellcolor{RankSecond}0.2199 & \cellcolor{RankFirst}\textbf{0.2652} & \textbf{0.2037} \\
Stone horse & \cellcolor{RankThird}0.2891 & 0.1248 & \cellcolor{RankFirst}0.2920 & \cellcolor{RankFirst}0.2387 & 0.2823 & \cellcolor{RankSecond}0.1706 & \cellcolor{RankSecond}\textbf{0.2906} & \cellcolor{RankThird}\textbf{0.1601} \\
\midrule
\rowcolor{gray!15} \multicolumn{9}{c}{\textbf{Part II: Overall Average}} \\
\midrule
Avg & \cellcolor{RankThird}0.2359 & \cellcolor{RankThird}0.0956 & 0.2251 & 0.0867 & \cellcolor{RankSecond}0.2396 & \cellcolor{RankSecond}0.1371 & \cellcolor{RankFirst}\textbf{0.2622} & \cellcolor{RankFirst}\textbf{0.1649} \\
\bottomrule
\end{tabular}%
}
\end{table*}

\subsection{Transport-Guided Optimization and Leakage Suppression}

In addition to providing view-wise semantic correspondences, the transport plan also carries another signal that is equally important for editing quality, namely, which Gaussians receive strong and consistent editing support and which ones, although visible in multiple views, are not effectively absorbed by the target semantics. This distinction is particularly important in language-driven 3D editing, because applying nearly uniform update strength to all candidate Gaussians would easily cause the editing signal to spread from the target region to surrounding background areas or adjacent structures, eventually leading to color leakage, texture drift, and unintended local modifications. To explicitly incorporate this variation in support strength into the optimization process, we aggregate, across views, the source mass and the actually transported mass of Gaussian $g_i$ as
\begin{equation}
\bar a_i = \sum_{v \in \mathcal{V}(i)} \omega_i^v a_i^v, \qquad
\bar w_i = \sum_{v \in \mathcal{V}(i)} \omega_i^v w_i^v.
\end{equation}
These two quantities correspond to the editable carrying capacity that the Gaussian possesses under the multi-view setting and the portion of that capacity that is actually absorbed by the target semantics. Based on this discrepancy, we further write the transport residual together with the corresponding edit gate as
\begin{equation}
r_i = \max(\bar a_i - \bar w_i, 0), \qquad
\gamma_i = \exp\Bigl(- \frac{r_i}{\tau_r + \delta} \Bigr),
\end{equation}
where $\tau_r>0$ is a temperature parameter. Under this design, Gaussians with strong and stable support are assigned larger values of $\gamma_i$, whereas Gaussians that are only weakly supported, or are more likely to lie outside the target region, are automatically assigned smaller gate values and are therefore more strongly suppressed during the subsequent update.

Built on this gating mechanism, the overall training objective is written as
\begin{equation}
\mathcal{L} = \lambda_{\mathrm{img}} \mathcal{L}_{\mathrm{img}} + \lambda_{\mathrm{sem}} \mathcal{L}_{\mathrm{sem}} + \lambda_{\mathrm{uot}} \mathcal{L}_{\mathrm{uot}} + \lambda_{\mathrm{leak}} \mathcal{L}_{\mathrm{leak}},
\end{equation}
\begin{equation}
\mathcal{L}_{\mathrm{img}} = \sum_{v=1}^{V} \lVert R(\mathcal{G}, \Pi_v) - \hat I_v \rVert_1, \qquad
\mathcal{L}_{\mathrm{sem}} = \sum_{i=1}^{N} \gamma_i \lVert s_i - \mathrm{sg}(z_i^{\star}) \rVert_2^2,
\end{equation}
\begin{equation}
\mathcal{L}_{\mathrm{uot}} = \sum_{v=1}^{V} \mathcal{T}_v(T_v^{\star}), \qquad
\mathcal{L}_{\mathrm{leak}} = \sum_{i=1}^{N} (1 - \gamma_i) \lVert c_i - c_i^{(0)} \rVert_1.
\end{equation}
This objective places image-level editing alignment, semantic convergence toward the canonical edit field, view-wise transport regularization, and leakage suppression outside the target region into a single optimization framework. The gate $\gamma_i$ allows Gaussians that are truly supported by the editing evidence to move more actively toward the canonical edit field $z_i^{\star}$, while Gaussians lacking sufficient support are constrained to remain closer to their original color state $c_i^{(0)}$. As a result, the model can respond to the textual editing instruction while maintaining more stable local control over the extent of modification. In the above formulation, $\mathrm{sg}(\cdot)$ denotes the stop-gradient operator, while densification, pruning, and practical solver details are deferred to Appendix.

\begin{figure*}[t]
    \centering
    \includegraphics[width=0.96\linewidth]{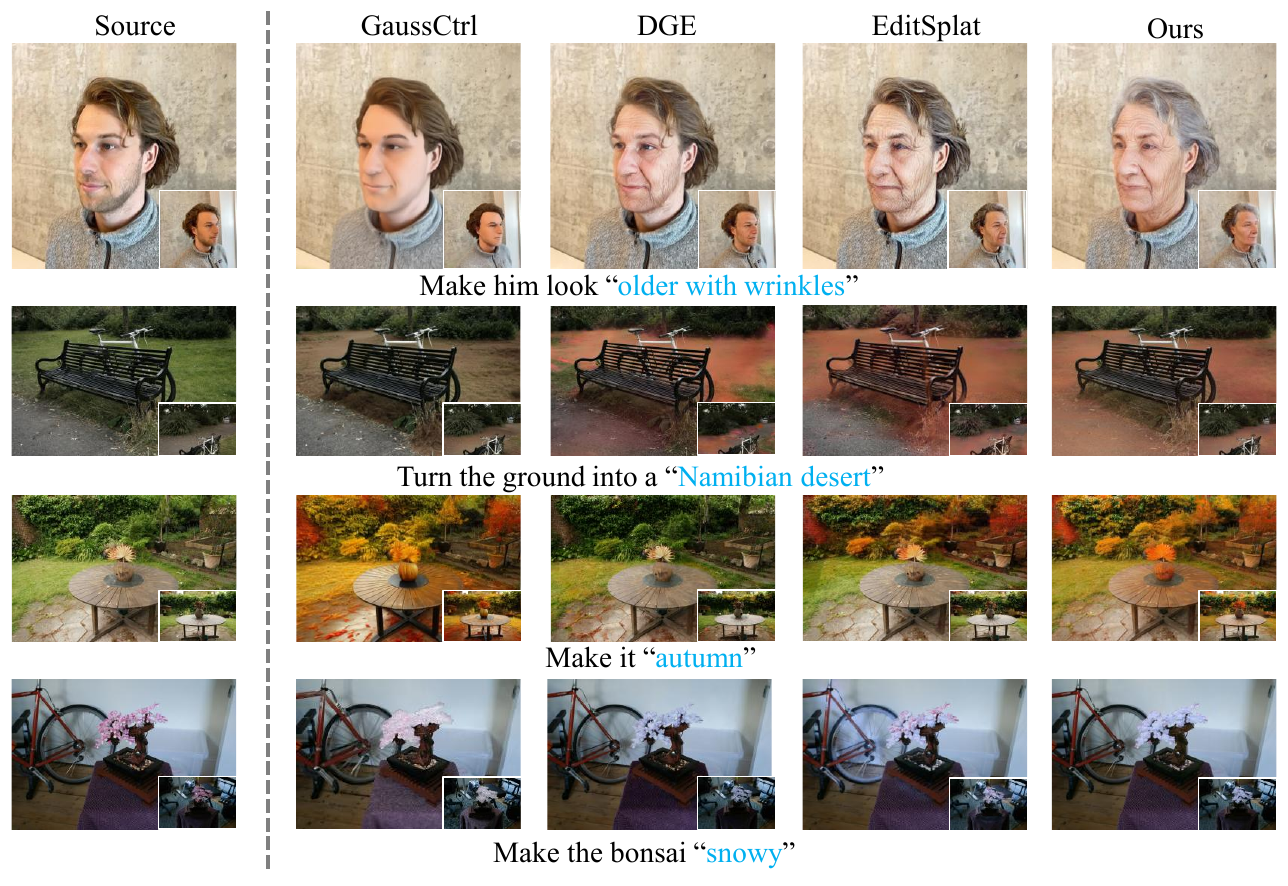}
    \caption{\textbf{Qualitative comparisons.} The leftmost column shows the source views, while the remaining columns present edited renderings produced by different methods under the same text instructions. An additional view is shown in the bottom-right corner of each result to examine cross-view consistency and local detail preservation.}
    \Description{Qualitative comparisons of different methods under the same text instructions, with source views in the leftmost column and additional views shown in the corner of each result for cross-view consistency comparison.}
    \label{fig:qualitative}
\end{figure*}

\subsection{Theoretical analysis}

Our theoretical analysis focuses on two main questions: whether the view-wise unbalanced semantic transport problem admits a well-defined and unique solution, and whether the canonical edit field obtained from multi-view fusion is stable. The former ensures that the transport correspondence remains deterministic when solved repeatedly across optimization views, while the latter shows that the shared semantic target is not overly sensitive to incomplete or noisy observations. 

\begin{proposition}
Assume that every entry of $a_v$ and $b_v$ is strictly positive and that $\varepsilon>0$, $\tau_s>0$, and $\tau_t>0$. Then the view-wise unbalanced semantic transport objective admits a unique minimizer on its feasible set. In other words, the view-wise transport operator is well defined.
\end{proposition}

\begin{theorem}
For a fixed Gaussian index $i$, if $\sum_{v\in\mathcal{V}(i)}\omega_i^v+\rho>0$, then the canonical edit field admits a unique minimizer given by
\begin{equation}
z_i^{\star}=
\frac{\sum_{v\in\mathcal{V}(i)}\omega_i^v y_i^v+\rho s_i}
{\sum_{v\in\mathcal{V}(i)}\omega_i^v+\rho}.
\label{eq:closedform}
\end{equation}
Moreover, under the same weights, if another set of observations yields $\tilde z_i^{\star}$, then the following stability bound holds:
\begin{equation}
\lVert z_i^{\star}-\tilde z_i^{\star}\rVert_2
\le
\frac{\sum_{v\in\mathcal{V}(i)}\omega_i^v\lVert y_i^v-\tilde y_i^v\rVert_2
+\rho \lVert s_i-\tilde s_i\rVert_2}
{\sum_{v\in\mathcal{V}(i)}\omega_i^v+\rho}.
\label{eq:stability}
\end{equation}
\end{theorem}

Theorem 1 shows that the canonical edit field is a confidence-weighted stable estimator: noise from any single view cannot be amplified beyond its normalized contribution. This property is especially important in language-driven 3D editing, where each view often provides only partial and noisy evidence for the target edit.

Overall, these results indicate that the view-wise transport module provides a deterministic intermediate semantic correspondence, while the canonical edit field further ensures stable cross-view fusion. More detailed derivations, variance analysis under noisy observations, and related corollaries are deferred to the appendix.

\section{Experiments}
\subsection{Experimental Setup and Baselines}

We adopt InstructPix2Pix (IP2P) as the 2D image editor and evaluate our method under the same benchmark protocol as EditSplat \cite{lee2025editsplat} to ensure a fair comparison. Specifically, our experiments are conducted on eight scenes, including four scenes from IN2N \cite{haque2023instruct}, one scene from BlendedMVS \cite{yao2020blendedmvs}, and three scenes from Mip-NeRF360 \cite{barron2022mip}. These scenes cover the datasets commonly used in prior language-driven 3D editing work while also including challenging large-scale real-world 360$^\circ$ scenes. All experiments are run on a single RTX A800 GPU. Under this hardware configuration, the full editing process for the \emph{Face} scene takes approximately 4 minutes.

For quantitative comparison, we use EditSplat, DGE, and GaussCtrl as the main baselines in the primary results table. Among them, EditSplat \cite{lee2025editsplat} is the most direct reference method, while DGE \cite{chen2024dge} and GaussCtrl \cite{wu2024gaussctrl} represent other representative directions in recent 3DGS editing research. For GaussCtrl, we follow its original setting and use the corresponding ControlNet-based editing system together with prompts in the style described by the method. Since GaussCtrl is sensitive to its textual input conditions, we additionally generate source-scene descriptions to reproduce its intended input format as closely as possible, thereby making the comparison more faithful and fair. Apart from such necessary adaptations, we keep the experimental protocol as consistent as possible with the publicly available settings of all baselines.

\begin{figure*}[t]
    \centering
    \includegraphics[width=0.96\linewidth]{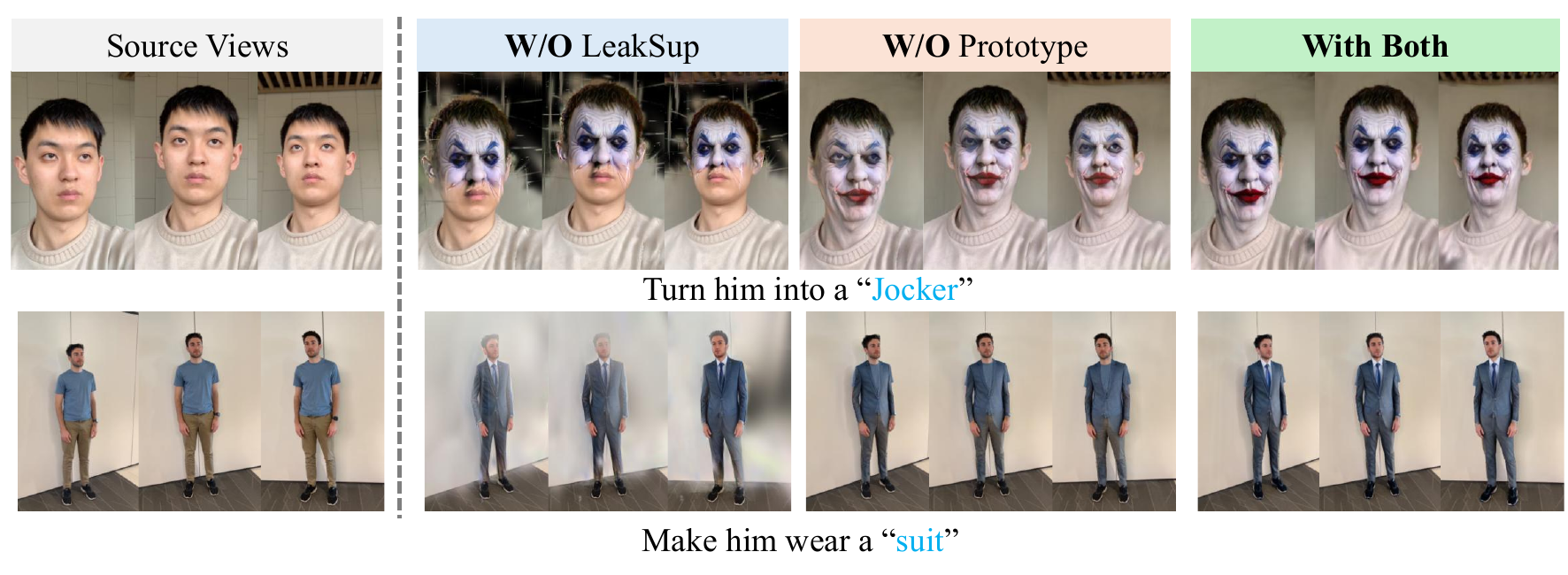}
    \caption{\textbf{Qualitative ablation results.} The leftmost column shows the source views, while the remaining columns compare the editing results of the variant without leakage suppression, the variant without prototype extraction, and the full model. The figure illustrates the role of prototype extraction in stabilizing local semantic evidence and the role of leakage suppression in reducing unintended edits outside the target region.}
    \Description{Qualitative ablation results comparing source views, a variant without leakage suppression, a variant without prototype extraction, and the full model.}
    \label{fig:ablation}
\end{figure*}

\subsection{Main quantitative results}
Table~\ref{tab:main_results} presents the main quantitative comparison results. We conduct a unified evaluation of all methods on 8 scenes and their corresponding text prompts. The table reports two metrics: CLIP-sim and CLIP-dir. CLIP-sim measures the overall semantic alignment between the edited results and the target text, while CLIP-dir evaluates whether the edits move along the intended semantic direction.

As shown in the table, our method achieves the best overall average performance on the benchmark, with CLIP-sim and CLIP-dir reaching 0.2622 and 0.1649, respectively. Compared with the strongest baseline, EditSplat, which obtains 0.2396 and 0.1371, our method improves these two metrics by 0.0226 and 0.0278, respectively. At the same time, our method also achieves better per-scene results on multiple scenes, including Bear, Bicycle, Bonsai, Face, Fangzhou, and Garden, indicating that the improvement does not come from only a few isolated examples, but remains stable across different types of editing scenarios. Compared with existing methods, our approach responds more accurately to text instructions and produces edits that are more consistent with the target semantics. This advantage is particularly evident in more challenging cases, such as non-rigid edits, material transformations, and local identity changes. Although simple multi-view fusion can produce visually similar results across viewpoints, it may still introduce inconsistent semantic updates over shared 3D regions. In contrast, our view-wise semantic transport together with the shared canonical 3D edit field provides a more unified target for 3D updates, thereby reducing semantic drift and yielding more reliable editing results.

\begin{table}[t]
\centering
\caption{\textbf{Ablation studies on core components.} We evaluate the impact of Transport-Guided Leakage Suppression (LeakSup) and Prompt-Aware Prototype extraction. The best, second, and third results are highlighted with \colorbox{RankFirst}{green}, \colorbox{RankSecond}{yellow}, and \colorbox{RankThird}{blue} backgrounds.}
\label{tab:ablation_results}
\resizebox{\linewidth}{!}{%
\begin{tabular}{@{}l | cc | cc@{}}
\toprule
\multirow{2}{*}{\textbf{Model Variant}} & \multicolumn{2}{c|}{\textbf{Components}} & \multicolumn{2}{c}{\textbf{Metrics}} \\
\cmidrule(lr){2-3} \cmidrule(lr){4-5}
& \textbf{LeakSup} & \textbf{Prototype} & \textbf{CLIP-sim}$\uparrow$ & \textbf{CLIP-dir}$\uparrow$ \\
\midrule
\rowcolor{gray!15} \multicolumn{5}{c}{\textbf{Part I: Ablation Configurations}} \\
\midrule
w/o LeakSup   & \xmark & \cmark & \cellcolor{RankSecond}0.2552 & \cellcolor{RankThird}0.1954 \\
w/o Prototype & \cmark & \xmark & \cellcolor{RankThird}0.2442  & \cellcolor{RankSecond}0.1989 \\
\midrule
\rowcolor{gray!15} \multicolumn{5}{c}{\textbf{Part II: Complete Architecture}} \\
\midrule
\textbf{Ours (Full)} & \textbf{\cmark} & \textbf{\cmark} & \cellcolor{RankFirst}\textbf{0.2652} & \cellcolor{RankFirst}\textbf{0.2037} \\
\bottomrule
\end{tabular}%
}
\end{table}

\subsection{Ablation Studies}

We first ablate \textit{prompt-aware prototype extraction} and \textit{transport-guided leakage suppression}, as these two modules respectively contribute to the stable modeling of local semantic evidence and the suppression of unintended edits outside the target region. As shown in Table~2, keeping either Prototype or LeakSup alone already brings noticeable improvements, while enabling both yields the best CLIP Score and CLIP Directional Score. This suggests that stable local prototype representations and leakage suppression play complementary roles in improving semantic alignment and editing-direction consistency. Qualitative results in Fig.~\ref{fig:ablation} further support this observation. Removing Prototype makes local semantic evidence more vulnerable to viewpoint variation and fine-grained noise, leading to less stable makeup intensity, placement, and detail in face editing. Removing LeakSup, in contrast, makes the edit more likely to spread beyond the intended target region, thereby weakening local control precision. By comparison, the full model maintains both more stable local semantic expressions and cleaner target-region updates. Since \textit{view-wise UOT} is more tightly coupled with the overall correspondence modeling process, its contribution is reflected not only within our framework but also in its transferability across different editing pipelines. We therefore further investigate UOT in the appendix by introducing it as a plug-and-play module into existing methods, while keeping the remaining settings as unchanged as possible, to validate its effectiveness and generalizability.

\subsection{Qualitative evaluation}
As shown in Fig.~\ref{fig:teaser} and Fig.~\ref{fig:qualitative}, we present representative qualitative results and conduct comparisons on several editing tasks, including facial aging, localized scene attribute transformation, seasonal appearance editing, and fine-grained object modification. Existing methods can often produce visually plausible results in a single view, but under multi-view settings they still tend to suffer from unstable local semantics, insufficient target modification, or interference with nearby regions. In the facial aging task, some baselines produce inconsistent wrinkle patterns, drifting facial textures, or overly smoothed structures. In the tasks of transforming the ground into a Namibian desert and changing the scene to an autumn style, some methods exhibit excessive color transfer, expanded editing regions, or insufficient preservation of local details. In the bonsai snow-scene editing example, non-target regions are also more easily affected. By contrast, our method more accurately confines the edit to the target region while preserving the original scene structure and geometric plausibility, producing results that are more natural, cleaner, and more consistent across views. The inset views in the bottom-right corners further show that our method maintains more stable local details and overall editing effects under viewpoint changes.

\section{Conclusion}
We presented TransSplat, a framework for language-driven 3D Gaussian editing that formulates the 2D-to-3D editing process as multi-view unbalanced semantic transport combined with canonical barycentric fusion. Unlike methods that rely only on cross-view appearance consistency, our approach further estimates cross-view correspondences between edited 2D evidence and shared 3D Gaussians. The resulting canonical edit field provides a unified target for 3D updates, while transport residuals help reduce edit leakage in weakly supported regions. Our analysis shows that the view-wise transport solution is well defined and that the canonical edit field remains stable under noisy and incomplete observations. Experimental results show that, compared with existing methods mainly designed for view consistency, our method improves both local editing accuracy and structural consistency. The study further suggests that improving multi-view fusion alone is not sufficient; reliably mapping view-dependent editing evidence to a shared 3D representation is also crucial for achieving robust multi-view editing.

\newpage
\bibliography{example_paper}
\bibliographystyle{icml2026}

\newpage
\appendix

\setcounter{tocdepth}{2}
\newcommand{\addmainsecentry}[3]{%
  \addtocontents{toc}{\protect\contentsline {section}{\protect\numberline {#1}#2}{#3}{tocstay}}}
\newcommand{\addmainsubsecentry}[3]{%
  \addtocontents{toc}{\protect\contentsline {subsection}{\protect\numberline {#1}#2}{#3}{tocstay}}}

\clearpage
\twocolumn[%
\begin{center}
{\Huge\sffamily\bfseries TransSplat: Unbalanced Semantic Transport for Language-Driven 3DGS Editing\par}
\medskip
{\Huge\sffamily\bfseries Supplementary Material\par}
\end{center}
\medskip
]

\section*{Contents}
{%
\renewcommand{\contentsname}{}%
\let\origaddcontentsline\addcontentsline%
\renewcommand{\addcontentsline}[3]{}%
\tableofcontents%
\let\addcontentsline\origaddcontentsline%
}

\section{Extended Method Details}

\subsection{Prototype Extraction}

For each optimization view $v$, the 2D editor produces an edited image $\hat I^v$, a text-relevant attention map $A^v \in \mathbb{R}^{H \times W}$, and an edited feature map $E^v \in \mathbb{R}^{H \times W \times d_e}$. To complement semantic evidence with local cues such as material, texture, and color, we further extract a pixel-level appearance feature map $\Phi^v \in \mathbb{R}^{H \times W \times d_a}$ from the same edited view. The feature map $E^v$ describes what the region is expected to become after editing, while $\Phi^v$ captures how that region appears locally. Rather than sending dense pixel-level evidence directly into the correspondence module, we first compress it into a compact set of structured prompt-aware prototypes.

The first step is to determine an effective support region from the attention map. Let $p$ denote a pixel location. We normalize the attention response as
\begin{equation}
\bar A^v(p)=\frac{A^v(p)}{\max_{q} A^v(q)+\epsilon_a},
\end{equation}
which places responses from different views on a comparable scale and makes the subsequent thresholding step more stable. If only the attention signal is used, the support region is defined as
\begin{equation}
\Omega^v=\{\,p \mid \bar A^v(p)\ge \delta_a\,\},
\end{equation}
so that only pixels with sufficiently strong text response are retained. When an external segmentation mask is available, we further restrict the support to
\begin{equation}
\Omega^v=\{\,p \mid \bar A^v(p)\ge \delta_a\,\}\cap M_{\mathrm{sam}}^v,
\end{equation}
with $M_{\mathrm{sam}}^v$ denoting the target mask of the edited region in view $v$. In practice, we additionally remove very small isolated connected components to suppress scattered noisy responses. After this stage, prototype extraction is concentrated on the most text-responsive local region instead of being dominated by large irrelevant background areas.

Once the support region $\Omega^v$ has been determined, we partition it into $M_v$ local regions by confidence-aware weighted clustering. Let $\xi(p)\in\mathbb{R}^2$ denote the 2D coordinate of pixel $p$. The clustering objective is
\begin{equation}
\min_{\{\Omega_m^v\}_{m=1}^{M_v},\,\{\mu_m^v\}_{m=1}^{M_v}}
\sum_{m=1}^{M_v}\sum_{p\in\Omega_m^v}\bar A^v(p)\,\|\xi(p)-\mu_m^v\|_2^2,
\end{equation}
and the resulting local regions satisfy
\begin{equation}
\Omega^v=\bigcup_{m=1}^{M_v}\Omega_m^v,
\qquad
\Omega_m^v \cap \Omega_n^v = \varnothing \;\; (m\neq n).
\end{equation}
Because pixels with larger attention values receive larger weights, the resulting regions naturally concentrate around locations that carry stronger editing evidence. This step reorganizes the dense support region into a small number of local fragments that already reflect textual relevance before any 2D--3D correspondence is computed.

For each local region $\Omega_m^v$, we construct a prototype
\begin{equation}
p_m^v=(u_m^v,e_m^v,b_m^v),
\end{equation}
whose three components encode the 2D location, the semantic descriptor, and the confidence mass of that region. We define the prototype position as the attention-weighted centroid
\begin{equation}
u_m^v=
\frac{\sum_{p\in\Omega_m^v}\bar A^v(p)\,\xi(p)}
{\sum_{p\in\Omega_m^v}\bar A^v(p)}.
\end{equation}
Compared with a plain geometric center, this centroid is biased toward pixels with stronger editing responses and is therefore more representative of the actual text-driven region.

The semantic descriptor is obtained by aggregating edited features inside the same local region:
\begin{equation}
e_m^v=
\frac{\sum_{p\in\Omega_m^v}\bar A^v(p)\,E^v(p)}
{\sum_{p\in\Omega_m^v}\bar A^v(p)},
\qquad
e_m^v \leftarrow \frac{e_m^v}{\|e_m^v\|_2+\epsilon_e}.
\end{equation}
The normalization improves the stability of the cosine similarity used later in the transport cost. We further define the prototype mass by the relative amount of accumulated attention inside the same region:
\begin{equation}
b_m^v=
\frac{\sum_{p\in\Omega_m^v}\bar A^v(p)}
{\sum_{p\in\Omega^v}\bar A^v(p)},
\qquad
\sum_{m=1}^{M_v} b_m^v=1.
\end{equation}
This quantity indicates how much reliable editing evidence is carried by the $m$-th prototype and simultaneously induces a normalized target mass distribution for the subsequent transport problem.

Semantic descriptors alone are often insufficient when two local regions are semantically related but visually very different. To incorporate local visual cues, we also define an appearance descriptor for each prototype:
\begin{equation}
\psi_m^v=
\frac{\sum_{p\in\Omega_m^v}\bar A^v(p)\,\Phi^v(p)}
{\sum_{p\in\Omega_m^v}\bar A^v(p)},
\qquad
\psi_m^v \leftarrow \frac{\psi_m^v}{\|\psi_m^v\|_2+\epsilon_\psi}.
\end{equation}
The descriptor $\psi_m^v$ captures local appearance attributes such as texture, material, and color. It does not replace semantic information; instead, it provides an additional visual cue that helps suppress mismatches between regions that are nearby in image space yet not truly aligned in local appearance.

After this stage, the dense editing evidence in view $v$ has been compressed into a compact set of prompt-aware prototypes. This substantially reduces the cost of subsequent correspondence modeling while preserving the location, semantic, confidence, and appearance information required for stable 2D--3D transport.

\subsection{View-wise Unbalanced Semantic Transport}

Once the prototypes have been extracted, the next question is how the local 2D editing evidence in the current view should be assigned to the visible 3D Gaussians. We treat the visible Gaussians as a source measure and the prototype set as a target measure, and solve a view-wise unbalanced semantic transport problem between them. For view $v$, let $\mathcal I_v$ denote the index set of visible Gaussians. Each Gaussian is represented as
\begin{equation}
g_i=(x_i,\Sigma_i,c_i,\alpha_i,s_i),
\end{equation}
with $x_i$ the center, $\Sigma_i$ the covariance, $c_i$ the color, $\alpha_i$ the opacity, and $s_i\in\mathbb{R}^{d_e}$ the semantic latent. In our formulation, $s_i$ stores the editable semantic content attached to the Gaussian, while explicit appearance remains represented by $c_i$ and $\alpha_i$.

We begin by defining the source and target measures. For each visible Gaussian, the source mass is written as
\begin{equation}
a_i^v=
\frac{\nu_i^v\,\alpha_i}
{\sum_{j\in\mathcal I_v}\nu_j^v\,\alpha_j},
\qquad i\in\mathcal I_v,
\end{equation}
where $\nu_i^v\in[0,1]$ denotes the visibility weight of Gaussian $i$ under view $v$. This definition reflects how strongly the Gaussian contributes to the rendered image in the current view, and therefore how much editing signal it is able to carry there. On the target side, we directly inherit the prototype confidence distribution:
\begin{equation}
b^v=(b_1^v,\dots,b_{M_v}^v),
\qquad
\sum_{m=1}^{M_v} b_m^v=1.
\end{equation}
Accordingly, the source side specifies which visible Gaussians are capable of absorbing edits, while the target side specifies how the local edited evidence is distributed across prototypes. In practice, zero-mass source or target entries are discarded before solving the transport problem.

We then define the transport cost between Gaussian $i$ and prototype $m$ in view $v$ as
\begin{equation}
C_{im}^v=
\lambda_{\mathrm{geo}}\|\Pi_v(x_i)-u_m^v\|_2^2
+\lambda_{\mathrm{sem}}\bigl(1-\cos(s_i,e_m^v)\bigr)
+\lambda_{\mathrm{app}}\bigl(1-\cos(\phi_i^v,\psi_m^v)\bigr),
\end{equation}
where $\Pi_v(x_i)$ denotes the 2D projection of Gaussian center $x_i$ in view $v$. The three terms account for image-plane locality, semantic discrepancy, and appearance discrepancy, respectively. As a result, the correspondence is not determined by geometry alone, but by the joint agreement of location, semantics, and local visual appearance.

To make the appearance term explicit, we define the Gaussian appearance descriptor $\phi_i^v$ by aggregating pixel-level appearance features over the projected footprint of Gaussian $i$ in the edited view. Let $\mathcal R_i^v$ denote the projected support region of Gaussian $i$, and let $\kappa_i^v(p)$ denote its normalized rendering contribution at pixel $p$. We compute
\begin{equation}
\phi_i^v=
\frac{\sum_{p\in\mathcal R_i^v}\kappa_i^v(p)\,\Phi^v(p)}
{\sum_{p\in\mathcal R_i^v}\kappa_i^v(p)},
\qquad
\phi_i^v \leftarrow \frac{\phi_i^v}{\|\phi_i^v\|_2+\epsilon_\phi}.
\end{equation}
Intuitively, $\phi_i^v$ describes how Gaussian $i$ appears in the current view, while $\psi_m^v$ describes the local appearance of the target edited region. Comparing the two helps suppress assignments that are spatially nearby but visually implausible.

Based on the above definitions, we solve the following view-wise unbalanced transport problem:
\begin{equation}
T_v^\star
=
\arg\min_{T_v\ge 0}
\langle C_v,T_v\rangle
+\varepsilon\,\mathrm{KL}(T_v\|a_v b_v^\top)
+\tau_s\,\mathrm{KL}(T_v\mathbf 1\|a_v)
+\tau_t\,\mathrm{KL}(T_v^\top\mathbf 1\|b_v),
\end{equation}
where $T_v\in\mathbb{R}_{+}^{|\mathcal I_v|\times M_v}$ is the transport plan, $\varepsilon$ is the entropic regularization coefficient, and $\tau_s,\tau_t>0$ control the source and target marginal relaxations. The unbalanced formulation is essential in our setting. Language-driven editing is rarely mass-preserving in a strict sense: local material changes, structural removal, and identity transformations may leave part of the source mass unmatched, while some target evidence may only be partially supported by the visible 3D structure. Allowing such mismatch makes the transport model better aligned with the actual editing process.

Once $T_v^\star$ is obtained, we first define the transported mass received by Gaussian $i$ in view $v$ as
\begin{equation}
w_i^v=\sum_{m=1}^{M_v} T_{im}^{v\star}.
\end{equation}
This quantity measures how strongly the Gaussian is supported by the edited evidence in the current view. We then define the corresponding view-induced semantic target as
\begin{equation}
y_i^v=
\frac{\sum_{m=1}^{M_v}T_{im}^{v\star} e_m^v}
{\sum_{m=1}^{M_v}T_{im}^{v\star}+\epsilon_y}.
\end{equation}
The variable $y_i^v$ specifies the semantic direction that Gaussian $g_i$ should follow under view $v$, while $w_i^v$ quantifies how reliable that suggestion is. Together, they form the output of the view-wise correspondence module and provide the inputs for the subsequent cross-view fusion stage.

\subsection{Canonical Barycentric Edit Field}

The view-wise targets obtained above are still incomplete and view-specific. A single view usually observes only part of the object, and the corresponding edited evidence may be noisy, ambiguous, or locally inconsistent with that of other views. Directly updating the 3D representation from each view in isolation would therefore be unstable. We instead fuse the view-induced targets into a unified canonical edit field shared across views.

For Gaussian $i$, let
\begin{equation}
\mathcal V(i)=\{\,v \mid i\in\mathcal I_v,\; w_i^v>0\,\}
\end{equation}
denote the set of views that provide valid transported evidence. We then define the confidence weight of view $v$ for Gaussian $i$ as
\begin{equation}
\omega_i^v=
\frac{w_i^v}{\sum_{v'\in\mathcal V(i)} w_i^{v'}},
\qquad
v\in\mathcal V(i).
\end{equation}
This normalization turns transported support into an adaptive fusion weight. Views that provide stronger and cleaner evidence contribute more, whereas weakly supported or noisy views are automatically downweighted.

The canonical semantic target $z_i^\star$ is defined by solving
\begin{equation}
z_i^\star
=
\arg\min_{z}
\sum_{v\in\mathcal V(i)}\omega_i^v\|z-y_i^v\|_2^2
+\rho\|z-s_i\|_2^2,
\end{equation}
where $\rho>0$ is a regularization coefficient that anchors the fused target to the current semantic latent $s_i$. The first term encourages agreement with the view-induced targets, while the second prevents excessive drift when multi-view evidence is sparse or unstable. This objective admits the closed-form solution
\begin{equation}
z_i^\star=
\frac{\sum_{v\in\mathcal V(i)}\omega_i^v y_i^v+\rho s_i}
{\sum_{v\in\mathcal V(i)}\omega_i^v+\rho}.
\end{equation}

It is worth emphasizing that this formulation is not a heavy global Wasserstein barycenter solver. Instead, it is a lightweight confidence-weighted regularized barycentric fusion performed directly in the latent semantic space. We still refer to the result as a canonical barycentric edit field because it recovers a shared cross-view semantic target, while the actual implementation remains simple, closed-form, and stable.

\begin{figure*}[t]
\centering
\includegraphics[width=0.72\textwidth]{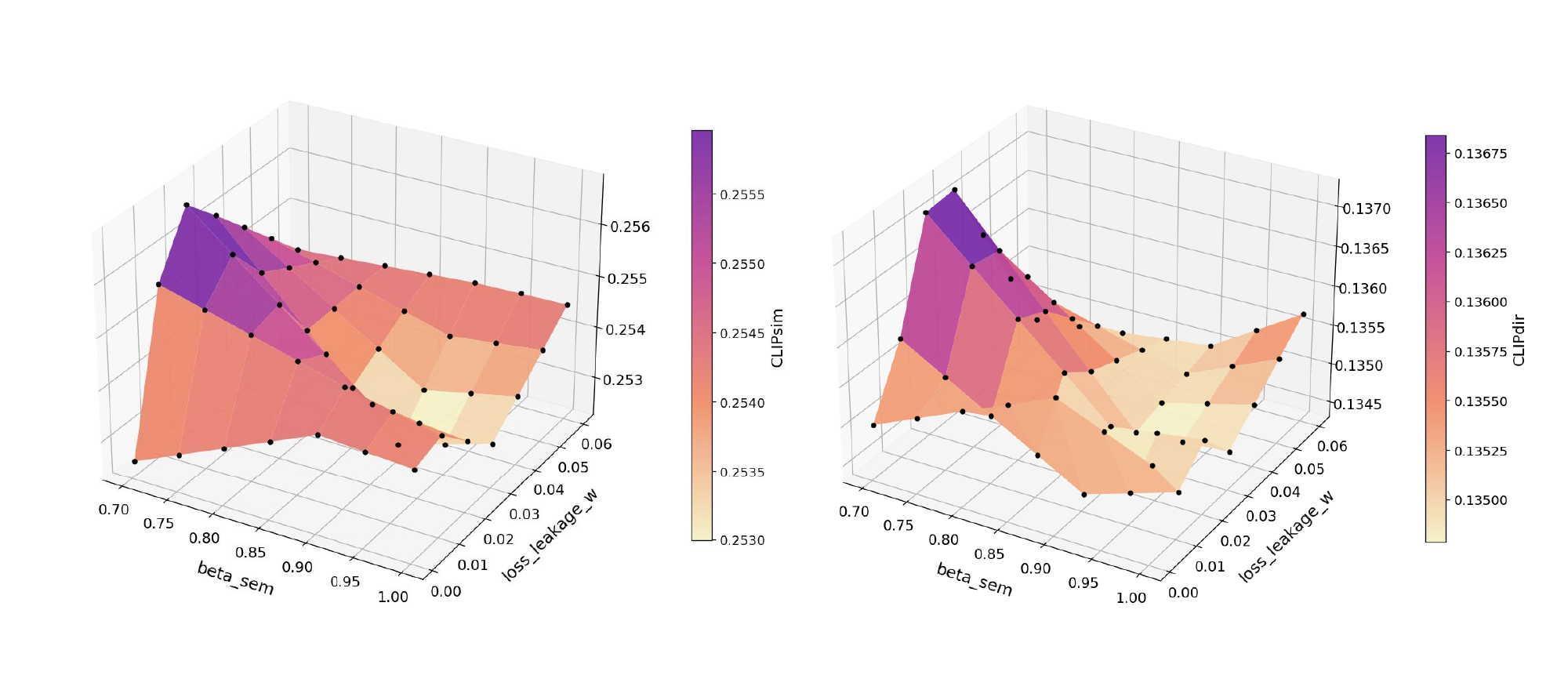}
\caption{Hyperparameter analysis. We show the influence of the semantic transport weight $\beta_{\mathrm{sem}}$ and the leakage suppression weight $\texttt{loss\_leakage\_w}$ under different parameter combinations on editing performance. The left and right plots report CLIP-sim and CLIP-dir, respectively.}
\Description{Representative editing results of TransSplat on pretrained 3D Gaussian Splatting scenes under text-guided instructions.}
\label{fig:hyperparam_appendix}
\end{figure*}

This construction offers two practical advantages. When multiple views provide consistent evidence for the same Gaussian, $z_i^\star$ naturally moves toward their shared semantic target. When the evidence is sparse, incomplete, or noisy, the anchor term $\rho\|z-s_i\|_2^2$ prevents the fused target from drifting too far away from the current semantic latent. The resulting canonical edit field is therefore more stable than a naive average over partially observed view-specific targets.

\subsection{Transport-Guided Leakage Suppression}

Even with view-wise transport and canonical fusion, uniformly applying the fused target $z_i^\star$ to all Gaussians may still lead to edit leakage. In local editing scenarios, some Gaussians are geometrically close to the target region but receive only weak or unreliable support from the edited evidence. Updating those Gaussians as aggressively as the strongly supported ones can cause the edit to spill into background regions or nearby structures that should remain unchanged.

To quantify this effect, we first define the residual of Gaussian $i$ in view $v$ as
\begin{equation}
r_i^v=
\Bigl[a_i^v-\sum_{m=1}^{M_v}T_{im}^{v\star}\Bigr]_+,
\end{equation}
where $[\cdot]_+$ denotes the positive part. This residual measures how much of the source mass remains unmatched by transported editing evidence in the current view. A large residual indicates that the Gaussian is visible and participates in rendering, yet is not genuinely supported by the target prototypes. We then aggregate this signal across views:
\begin{equation}
r_i=
\sum_{v\in\mathcal V(i)}\omega_i^v r_i^v.
\end{equation}
Because the aggregation uses the same normalized weights as the canonical fusion stage, stronger and cleaner views contribute more to the final residual estimate.

Based on the aggregated residual, we define an editing gate
\begin{equation}
\gamma_i=\exp\!\left(-\frac{r_i}{\tau_r}\right),
\qquad
\gamma_i\in(0,1],
\end{equation}
where $\tau_r$ is a temperature parameter. Small residuals keep $\gamma_i$ close to one, meaning that the Gaussian is strongly supported by transported evidence. Large residuals reduce the gate value and make the corresponding update more conservative.

Using this gate, we no longer force all Gaussians to align to the canonical field with the same strength. Instead, the semantic update is modulated by transport support, and a simple gated target can be written as
\begin{equation}
\tilde s_i=\gamma_i z_i^\star + (1-\gamma_i)s_i.
\end{equation}
Under this formulation, strongly supported Gaussians move closer to the canonical edit field, whereas weakly supported ones remain closer to their original semantic state.

We also impose a leakage suppression loss on appearance parameters. Let $c_i^{(0)}$ denote the original color of Gaussian $i$ before editing begins. We define
\begin{equation}
L_{\mathrm{leak}}
=
\sum_i (1-\gamma_i)\,\|c_i-c_i^{(0)}\|_2^2.
\end{equation}
The term $c_i^{(0)}$ explicitly denotes the pre-edit color parameter rather than any intermediate value during optimization. This loss keeps weakly supported Gaussians close to their original appearance and reduces undesired modifications outside the target region.

At the same time, the semantic alignment loss is written as
\begin{equation}
L_{\mathrm{sem}}
=
\sum_i \gamma_i\,\|s_i-z_i^\star\|_2^2.
\end{equation}
These two losses are controlled by the same transport-guided confidence mechanism. Regions with strong support are encouraged to follow the canonical edit field more aggressively, whereas weakly supported ones are explicitly regularized toward preservation.

Finally, we clarify which Gaussian attributes are affected by the proposed module. The transport-guided canonical field acts directly on the semantic latent $s_i$ and influences the color $c_i$ and opacity $\alpha_i$ indirectly through rendering-based supervision. By contrast, the Gaussian center $x_i$ and covariance $\Sigma_i$ are not directly constrained by the transport objective; they remain governed mainly by the original 3DGS optimization process together with densification and pruning rules. In this sense, the proposed method provides local editing guidance at the semantic and appearance levels rather than aggressively rewriting the scene geometry itself.

Overall, transport-guided leakage suppression completes the full framework. The transport plan does not merely establish semantic correspondence; it also quantifies how strongly each Gaussian should be edited. This makes the update rule more interpretable and more robust in the presence of partial or noisy multi-view editing evidence.

\section{Practical Optimization Details}

\subsection{Detailed Experimental Setup}

\noindent\textbf{3DGS stage.}
TransSplat follows the same basic experimental setup as EditSplat. For each scene, we use a pretrained 3D Gaussian Splatting representation obtained after 30,000 iterations as the source scene, and perform the subsequent language-driven 3D editing on top of it. During the 3DGS training stage, we use the Adam optimizer, with the learning rate kept the same as in the original 3DGS implementation. Densification and pruning are enabled for all scenes.

\noindent\textbf{Editing stage.}
For the 2D image editor, we use InstructPix2Pix (IP2P) from the Diffusers library. In practice, we adopt the DDIM scheduler with 20 sampling steps, and sample the noise level from the timestep range $t \in [0.70,\,0.98]$. For each view, we extract 32 local prototypes and introduce a 16-dimensional semantic feature for each Gaussian. By default, we enable SAM constraints to shrink the local candidate region, which improves the stability of prototype extraction and local correspondence. For the view-wise UOT solver, we use 30 Sinkhorn iterations and set the transport top-$k$ to 128 in order to control the computational cost. For cross-view fusion, we adopt a lightweight barycentric EMA update. In practice, we first compute the closed-form canonical target in Eq.~(22), and then use an EMA-style update to smoothly apply it during optimization.

\subsection{Evaluation Metrics}

We follow the evaluation protocol suggested by MipNeRF360, using a train/test split in which every eighth image is held out for testing. For quantitative evaluation, we adopt the same two CLIP-based metrics used in DGE, namely CLIP directional similarity and CLIP image-text similarity. These metrics evaluate whether the edited result changes along the desired semantic direction and how well the final result aligns with the target text, respectively. We additionally conduct a user study reported in Appendix D.3, where human evaluators rate the edited results in terms of Target Semantics Alignment, Background Preservation, and Overall Coherence. For GaussCtrl, since its original method uses ControlNet as the 2D image editor, we keep its original setting unchanged and set the guidance scale to 5 for fair comparison.

\subsection{Hyperparameter Analysis}

To further analyze the key design choices of our method, we conduct an additional study on two core hyperparameters: $\beta_{\mathrm{sem}}$, which controls the weight of the semantic cost term in UOT, and $\texttt{loss\_leakage\_w}$, which controls the strength of the background leakage suppression term. The former determines the relative contribution of semantic consistency in the transport cost, while the latter controls how strongly non-target regions are constrained during optimization. Figure~\ref{fig:hyperparam_appendix} reports the resulting CLIP-sim and CLIP-dir under different combinations of these two hyperparameters.

As shown in Figure~1, both $\beta_{\mathrm{sem}}$ and $\texttt{loss\_leakage\_w}$ exhibit a clear effective range, and neither larger values nor smaller values consistently lead to better performance. When $\beta_{\mathrm{sem}}$ is too small, the model does not make sufficient use of semantic information, making it more difficult for the target semantics specified by the text prompt to be consistently propagated through cross-view transport. In contrast, when $\beta_{\mathrm{sem}}$ becomes too large, the two metrics no longer improve and may even decline, suggesting that overly strong semantic constraints can disrupt the balance among semantic, geometric, and appearance cues. In comparison, a moderate $\beta_{\mathrm{sem}}$ generally yields better overall performance.

A similar trend can be observed for $\texttt{loss\_leakage\_w}$. When this weight is too small, the suppression on non-target regions is insufficient, which limits the overall semantic alignment quality. On the other hand, when $\texttt{loss\_leakage\_w}$ becomes too large, the optimization becomes overly conservative, which is also unfavorable for achieving high CLIP-sim and CLIP-dir scores. Overall, a moderate $\texttt{loss\_leakage\_w}$ provides a better trade-off between target-region editing and background preservation. This observation also indicates that the proposed leakage suppression mechanism needs to be properly balanced with the semantic transport term in order to achieve the best performance.

\section{Proofs and Further Theoretical Analysis}

\subsection{Proof of Proposition 1}

We prove that the view-wise unbalanced semantic transport objective admits a unique minimizer. Consider the feasible set
\begin{equation}
\mathcal{D}_v=\{T\in\mathbb{R}_{+}^{|\mathcal{I}_v|\times M_v}\}.
\end{equation}
This set is closed and convex. The proof proceeds by establishing two properties of $\mathcal{T}_v$: strict convexity on $\mathcal{D}_v$ and existence of a minimizer on $\mathcal{D}_v$.

We first examine convexity. The linear term $\langle C_v,T\rangle$ is convex in $T$. The two marginal relaxation terms,
\[
T\mapsto \mathrm{KL}(T\mathbf{1}\|a_v)
\quad\text{and}\quad
T\mapsto \mathrm{KL}(T^\top\mathbf{1}\|b_v),
\]
are also convex, since the maps $T\mapsto T\mathbf{1}$ and $T\mapsto T^\top\mathbf{1}$ are linear and the generalized Kullback--Leibler divergence is convex on the nonnegative orthant. The key term is
\[
T\mapsto \mathrm{KL}(T\|a_v b_v^\top).
\]
Under the assumptions of the proposition, every entry of $a_v$ and $b_v$ is strictly positive, so every entry of $a_v b_v^\top$ is strictly positive as well. As a result, $\mathrm{KL}(T\|a_v b_v^\top)$ is finite on $\mathcal{D}_v$ and strictly convex in $T$. Since its coefficient satisfies $\varepsilon>0$, the full objective $\mathcal{T}_v$ is strictly convex on $\mathcal{D}_v$.

We next show that the minimum is attained. The generalized Kullback--Leibler divergence satisfies
\[
x\log x-x+1\rightarrow +\infty
\qquad\text{as }x\rightarrow +\infty.
\]
Consequently, $\mathrm{KL}(T\|a_v b_v^\top)\rightarrow +\infty$ whenever $\|T\|_F\rightarrow +\infty$. The remaining terms of $\mathcal{T}_v$ are bounded below on $\mathcal{D}_v$, so the full objective is coercive on $\mathcal{D}_v$. In addition, $\mathcal{T}_v$ is lower semicontinuous on $\mathcal{D}_v$. Since $\mathcal{D}_v$ is closed, a coercive lower semicontinuous objective must attain its minimum on this set.

At this point, existence follows from coercivity and lower semicontinuity, while uniqueness follows from strict convexity. Therefore, $\mathcal{T}_v$ admits a unique minimizer on $\mathcal{D}_v$. This completes the proof.

\subsection{Proof of Theorem 1}

For a fixed Gaussian index $i$, consider the objective
\begin{equation}
\mathcal{B}_i(z)=\sum_{v\in\mathcal{V}(i)}\omega_i^v\|z-y_i^v\|_2^2+\rho\|z-s_i\|_2^2.
\end{equation}
The theorem has two parts. The first establishes the closed-form solution. The second derives a stability bound under perturbations of the view-wise targets and the current semantic latent.

We begin with uniqueness. Since $\mathcal{B}_i(z)$ is quadratic in $z$, its Hessian is given by
\begin{equation}
\nabla^2\mathcal{B}_i(z)=2\Bigl(\sum_{v\in\mathcal{V}(i)}\omega_i^v+\rho\Bigr)I_d.
\end{equation}
Because $\sum_{v\in\mathcal{V}(i)}\omega_i^v=1$ and $\rho>0$, the quantity $\sum_{v\in\mathcal{V}(i)}\omega_i^v+\rho$ is strictly positive, so the Hessian is positive definite. This shows that $\mathcal{B}_i$ is strongly convex and therefore admits a unique minimizer.

The closed-form expression follows by setting the gradient to zero:
\begin{equation}
0=2\sum_{v\in\mathcal{V}(i)}\omega_i^v(z-y_i^v)+2\rho(z-s_i).
\end{equation}
Collecting the terms involving $z$ gives
\begin{equation}
\Bigl(\sum_{v\in\mathcal{V}(i)}\omega_i^v+\rho\Bigr)z
=
\sum_{v\in\mathcal{V}(i)}\omega_i^v y_i^v+\rho s_i.
\end{equation}
Dividing both sides by $\sum_{v\in\mathcal{V}(i)}\omega_i^v+\rho$ yields
\begin{equation}
z_i^\star=
\frac{\sum_{v\in\mathcal{V}(i)}\omega_i^v y_i^v+\rho s_i}
{\sum_{v\in\mathcal{V}(i)}\omega_i^v+\rho}.
\end{equation}
This proves the closed-form formula in the main text.

We now turn to stability. Let $\tilde y_i^v$ and $\bar s_i$ be perturbed inputs, and let the corresponding solution be
\begin{equation}
\tilde z_i^\star=
\frac{\sum_{v\in\mathcal{V}(i)}\omega_i^v \tilde y_i^v+\rho \bar s_i}
{\sum_{v\in\mathcal{V}(i)}\omega_i^v+\rho}.
\end{equation}
Subtracting the two expressions gives
\begin{equation}
z_i^\star-\tilde z_i^\star
=
\frac{\sum_{v\in\mathcal{V}(i)}\omega_i^v(y_i^v-\tilde y_i^v)+\rho(s_i-\bar s_i)}
{\sum_{v\in\mathcal{V}(i)}\omega_i^v+\rho}.
\end{equation}
Taking the Euclidean norm and applying the triangle inequality leads to
\begin{equation}
\|z_i^\star-\tilde z_i^\star\|_2
\le
\frac{\sum_{v\in\mathcal{V}(i)}\omega_i^v\|y_i^v-\tilde y_i^v\|_2+\rho\|s_i-\bar s_i\|_2}
{\sum_{v\in\mathcal{V}(i)}\omega_i^v+\rho}.
\end{equation}
This is exactly the desired perturbation bound. It shows that the canonical edit field depends continuously on the per-view semantic targets and on the current latent state, which completes the proof.

\subsection{Variance and Stability Analysis under Noisy Observations}

We now analyze the statistical behavior of the canonical edit field when the view-wise semantic targets are noisy. This analysis clarifies why the cross-view fusion step improves stability over direct per-view updates.

Assume that the semantic target from each valid view can be decomposed as
\begin{equation}
y_i^v=y_i^\dagger+\xi_i^v,
\end{equation}
in which $y_i^\dagger$ denotes the unknown ground-truth canonical semantic target and $\xi_i^v$ denotes the observation noise associated with view $v$. We further assume that the noise terms are zero mean,
\begin{equation}
\mathbb{E}[\xi_i^v]=0,
\end{equation}
and mutually independent across views.

We first consider the case $\rho=0$. The closed-form solution from Theorem~1 then reduces to
\begin{equation}
z_i^\star=\sum_{v\in\mathcal{V}(i)}\omega_i^v y_i^v,
\end{equation}
together with the normalization condition
\begin{equation}
\sum_{v\in\mathcal{V}(i)}\omega_i^v=1.
\end{equation}
Substituting the noisy observation model gives
\begin{equation}
z_i^\star
=
\sum_{v\in\mathcal{V}(i)}\omega_i^v(y_i^\dagger+\xi_i^v)
=
y_i^\dagger+\sum_{v\in\mathcal{V}(i)}\omega_i^v\xi_i^v.
\end{equation}
Taking expectations immediately yields
\begin{equation}
\mathbb{E}[z_i^\star]
=
y_i^\dagger+\sum_{v\in\mathcal{V}(i)}\omega_i^v\mathbb{E}[\xi_i^v]
=
y_i^\dagger.
\end{equation}
Thus, in the absence of the anchor term, the canonical edit field is an unbiased estimator of the true semantic target.

\begin{figure}[!t]
\centering
\includegraphics[width=\linewidth]{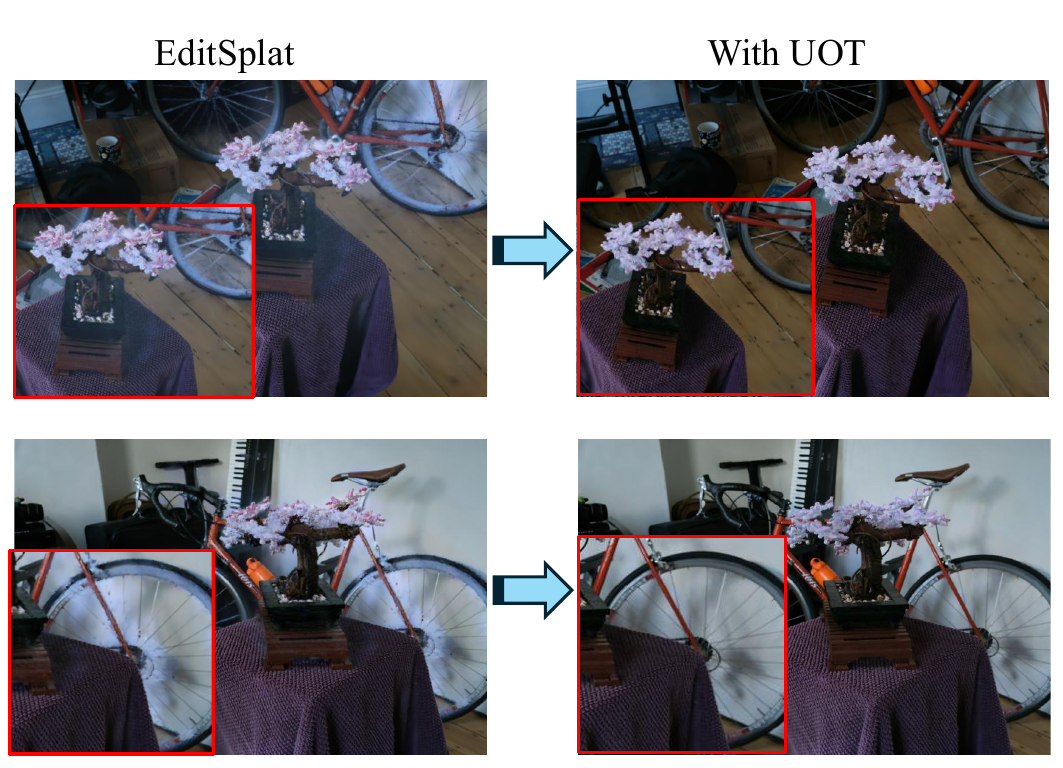}
\caption{Qualitative comparison between the original EditSplat and EditSplat augmented with the proposed view-wise UOT module.}
\Description{Qualitative comparison between EditSplat and EditSplat enhanced with the proposed view-wise UOT module.}
\label{fig:uot_appendix}
\end{figure}

The next step is to examine its second moment. From the previous identity,
\begin{equation}
z_i^\star-y_i^\dagger=\sum_{v\in\mathcal{V}(i)}\omega_i^v\xi_i^v.
\end{equation}
Consequently,
\begin{equation}
\mathbb{E}\bigl[\|z_i^\star-y_i^\dagger\|_2^2\bigr]
=
\mathbb{E}\Bigl[\Bigl\|\sum_{v\in\mathcal{V}(i)}\omega_i^v\xi_i^v\Bigr\|_2^2\Bigr].
\end{equation}
Expanding the squared norm produces both self terms and cross terms. Independence and zero mean remove the cross terms in expectation, leaving
\begin{equation}
\mathbb{E}\bigl[\|z_i^\star-y_i^\dagger\|_2^2\bigr]
=
\sum_{v\in\mathcal{V}(i)}(\omega_i^v)^2\,
\mathbb{E}\bigl[\|\xi_i^v\|_2^2\bigr].
\end{equation}
This identity already shows the variance-reduction effect of multi-view fusion: the final error is controlled by a weighted average of the per-view noise levels rather than by any single view alone.

A particularly transparent form appears when all visible views share the same second moment,
\begin{equation}
\mathbb{E}\bigl[\|\xi_i^v\|_2^2\bigr]=\sigma_i^2,
\end{equation}
and the weights are uniform,
\begin{equation}
\omega_i^v=\frac{1}{|\mathcal{V}(i)|}.
\end{equation}
Substituting these expressions gives
\begin{equation}
\mathbb{E}\bigl[\|z_i^\star-y_i^\dagger\|_2^2\bigr]
=
|\mathcal{V}(i)|\Bigl(\frac{1}{|\mathcal{V}(i)|}\Bigr)^2\sigma_i^2
=
\frac{\sigma_i^2}{|\mathcal{V}(i)|}.
\end{equation}
The mean-squared error therefore decreases at the rate $1/|\mathcal{V}(i)|$ as the number of informative views grows.

This analysis also helps explain the role of the anchor term. Once $\rho>0$, the fused target is no longer strictly unbiased, since the term $\rho s_i$ introduces a bias toward the current semantic latent. At the same time, this anchor further suppresses variance and makes the fused target more conservative. The parameter $\rho$ thus governs a bias--variance tradeoff: larger values improve stability, while smaller values allow the fusion result to follow the multi-view evidence more closely.

Taken together, these results provide a statistical interpretation of the canonical fusion step. The shared edit field is not only well defined and perturbation-stable, but also naturally reduces view-specific noise under standard assumptions. This is precisely why it offers a more reliable update target than direct per-view optimization.

\section{Additional Experiments}

\subsection{UOT Ablation on EditSplat}

As shown in Fig.~\ref{fig:uot_appendix}, when the proposed view-wise UOT module is integrated into EditSplat as a plug-and-play component while keeping the remaining settings unchanged as much as possible, the highlighted local regions exhibit clearer target boundaries, more coherent local structures, and more stable cross-view semantic correspondence. Compared with the original EditSplat, introducing UOT leads to more consistent updates of the same edited target across different viewpoints, reducing local ambiguity and instability near the boundary between the target region and the surrounding background. This result suggests that explicit view-wise semantic transport is effective not only within the full TransSplat framework, but also as an independent module for improving local editing precision and cross-view stability in existing 2D-to-3D editing pipelines.

\subsection{Additional Qualitative Results}

Beyond the representative examples shown in the main paper, we provide additional scene editing results under a larger set of prompts in the appendix, as shown in Fig.~\ref{fig:more_qualitative_appendix_1} and Fig.~\ref{fig:more_qualitative_appendix_2}. These supplementary examples cover several types of text-driven edits, including material changes, attribute modifications, and local semantic substitutions. Through these additional visualizations, we aim to show that our method not only produces strong results on a small number of selected examples, but also maintains stable local control and cross-view consistency under a broader range of prompts. Compared with the limited examples in the main paper, the supplementary results in Fig.~\ref{fig:more_qualitative_appendix_1} and Fig.~\ref{fig:more_qualitative_appendix_2} more clearly demonstrate the generalization ability of our method under richer text instructions.

\subsection{User Study}
\begin{figure*}[p]
\centering
\includegraphics[width=\textwidth,height=0.9\textheight,keepaspectratio]{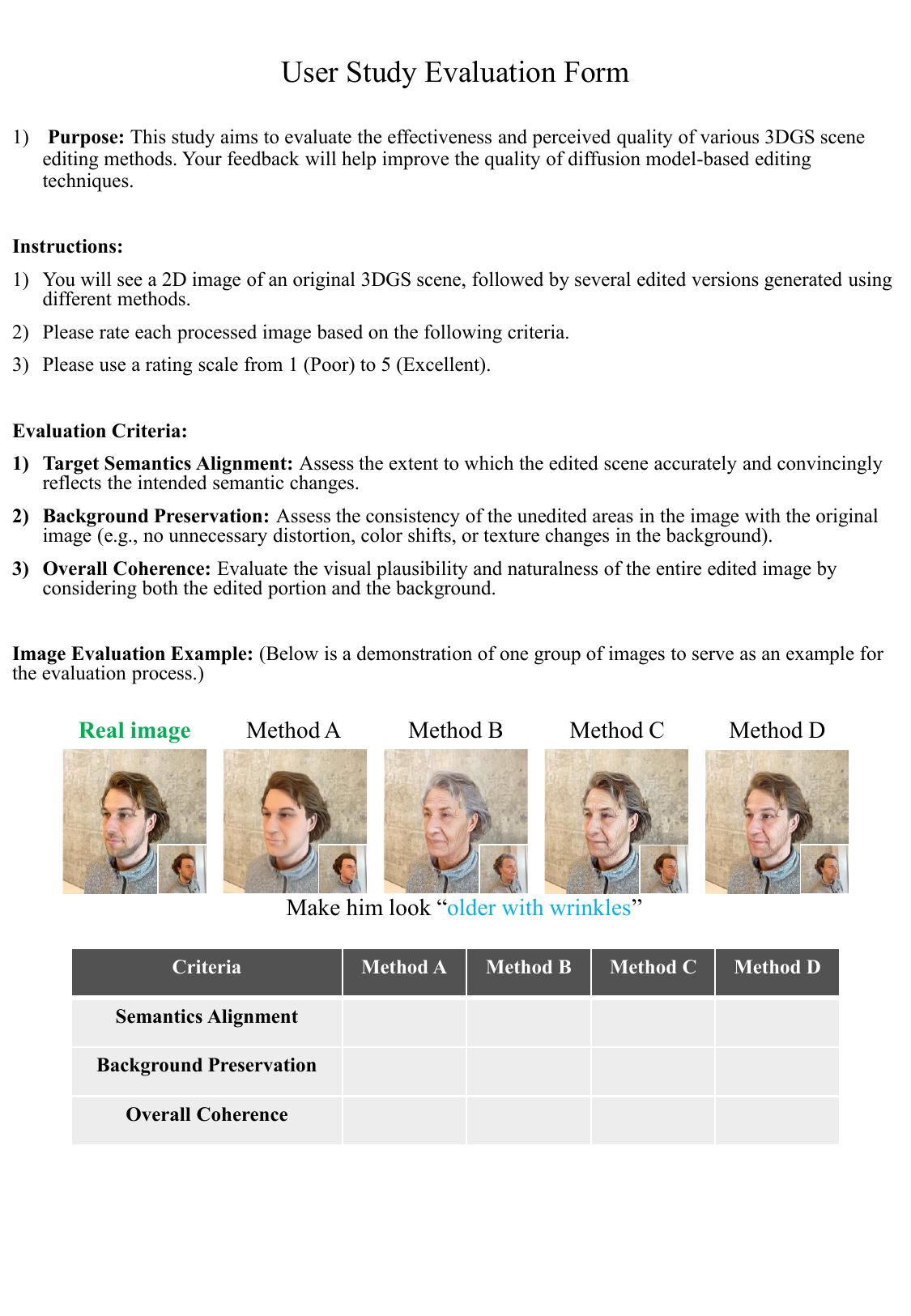}
\caption{User Study Form Example.}
\Description{Additional representative editing results of TransSplat on pretrained 3D Gaussian Splatting scenes under text-guided instructions.}
\label{fig:more_qualitative_appendix_3}
\end{figure*}

We further conduct a user study to complement the CLIP-based evaluation with human judgments. Specifically, we recruit 100 participants to subjectively assess the edited results produced by different methods. For each evaluation case, the participants are presented with the original image, the text instruction, and the edited results generated by all compared methods, and are asked to rate each result on a 1--5 scale according to three criteria: Target Semantics Alignment, Background Preservation, and Overall Coherence. Among them, Target Semantics Alignment evaluates whether the edited result accurately reflects the semantic change described by the text prompt; Background Preservation measures whether the non-edited regions remain consistent with the original image while minimizing artifacts, distortions, or unnecessary texture changes; Overall Coherence jointly assesses the visual plausibility and naturalness of the entire edited image by considering both the edited region and the preserved background. The evaluation form used in our user study is shown in Fig.~\ref{fig:more_qualitative_appendix_3}. We average the ratings over all participants and all evaluation cases, and report the resulting averages as the final user study results. The overall user study score is computed by averaging the ratings over all three criteria, all participants, and all evaluation cases.

The user study results are consistent with the quantitative findings in the main paper and show a clear human preference for our method. In particular, our method achieves the highest overall average score of 4.32, outperforming EditSplat (4.08), GaussCtrl (3.94), and DGE (3.79). A closer examination shows that this advantage mainly comes from Target Semantics Alignment and Overall Coherence. On Target Semantics Alignment, our method achieves an average score of 4.41, compared with 4.12 for EditSplat, 3.86 for GaussCtrl, and 3.72 for DGE, indicating that our method more accurately captures the intended text-driven semantic change and better matches user expectations. A similar trend is observed in Overall Coherence, where our method obtains 4.30, while EditSplat, GaussCtrl, and DGE achieve 4.02, 3.91, and 3.74, respectively. This suggests that our method produces edited results that are judged to be more visually natural and globally consistent.

In contrast, the gap among different methods is relatively smaller on Background Preservation. Our method obtains 4.25 on this criterion, while EditSplat, GaussCtrl, and DGE achieve 4.11, 4.06, and 3.92, respectively. This result suggests that most methods are already reasonably competitive in preserving non-target regions, whereas the main perceptual advantage of our method lies more in whether the edited semantics are accurate and whether the final results remain visually coherent and natural. Overall, the user study further demonstrates that the proposed view-wise semantic transport, cross-view canonical fusion, and transport-guided leakage suppression not only improve automatic metrics, but also produce edited scenes that are more preferred by human evaluators.

\clearpage
\begin{table*}[!t]
\caption{Notation summary. Main symbols used throughout the paper.}
\label{tab:SymbolsTable}
\begin{tabular}{@{}>{\raggedright\arraybackslash}p{0.31\textwidth} >{\raggedright\arraybackslash}p{0.64\textwidth}@{}}
\toprule
\textbf{Symbol} & \textbf{Description} \\
\midrule
$v, i, m, p$ & Optimization view index, Gaussian index, prototype index, and pixel location. \\
$H, W, d_e, d_a$ & Image height, image width, semantic feature dimension, and appearance feature dimension. \\
$\hat I^v, A^v, \bar A^v$ & Edited image in view $v$, raw text-relevant attention map, and its normalized version. \\
$E^v, \Phi^v, \xi(p)$ & Edited semantic feature map, pixel-level appearance feature map, and 2D image coordinate of pixel $p$. \\
$\Omega^v, M_{\mathrm{sam}}^v, \Omega_m^v, \mu_m^v, M_v$ & Prototype-extraction support region, optional SAM mask, the $m$-th local region, its clustering center, and the number of prototypes in view $v$. \\
$p_m^v=(u_m^v,e_m^v,b_m^v), \psi_m^v$ & Prompt-aware prototype with position centroid $u_m^v$, semantic descriptor $e_m^v$, confidence mass $b_m^v$, and appearance descriptor $\psi_m^v$. \\
$g_i=(x_i,\Sigma_i,c_i,\alpha_i,s_i), c_i^{(0)}$ & The $i$-th 3D Gaussian with center, covariance, color, opacity, semantic latent, and its original pre-edit color. \\
$\mathcal I_v, \nu_i^v, a_i^v, b^v$ & Visible Gaussian index set, visibility weight, source mass of Gaussian $i$, and prototype target-mass distribution in view $v$. \\
$C_{im}^v, \lambda_{\mathrm{geo}}, \lambda_{\mathrm{sem}}, \lambda_{\mathrm{app}}, \Pi_v(x_i)$ & View-wise transport cost, geometric/semantic/appearance weights, and the 2D projection of Gaussian center $x_i$ under view $v$. \\
$\mathcal R_i^v, \kappa_i^v(p), \phi_i^v, \psi_m^v$ & Projected support region of Gaussian $i$, its normalized rendering contribution at pixel $p$, and the Gaussian/prototype appearance descriptors. \\
$T_v, T_v^\star, \varepsilon, \tau_s, \tau_t, \mathrm{KL}(\cdot\|\cdot)$ & View-wise transport plan, optimal transport solution, entropic regularization coefficient, source/target marginal relaxation coefficients, and generalized KL divergence. \\
$w_i^v, y_i^v, \mathcal V(i), \omega_i^v$ & Transported support received by Gaussian $i$, the corresponding view-induced semantic target, the set of valid views, and the normalized fusion weight in view $v$, with $\sum_{v\in\mathcal V(i)} \omega_i^v = 1$. \\
$z_i^\star, \rho, \mathcal B_i(z)$ & Canonical fused semantic target of Gaussian $i$, anchor regularization coefficient, and canonical barycentric fusion objective. \\
$r_i^v, r_i, \tau_r, \gamma_i, \tilde s_i$ & Per-view residual, aggregated residual, residual temperature, editing gate, and gated semantic target. \\
$L_{\mathrm{sem}}, L_{\mathrm{leak}}$ & Semantic alignment loss and leakage suppression loss. \\
$\delta_a, \epsilon_a, \epsilon_e, \epsilon_\phi, \epsilon_\psi, \epsilon_y$ & Attention threshold and numerical stabilization constants used in normalization, transport, and fusion. \\
$\beta_{\mathrm{sem}}, \texttt{loss\_leakage\_w}$ & Hyperparameters used in the appendix hyperparameter analysis. \\
$\mathcal D_v, \mathcal T_v, \tilde y_i^v, \bar s_i, \tilde z_i^\star$ & Feasible set of transport plans, view-wise UOT objective, perturbed view-wise targets, perturbed latent state, and the corresponding perturbed fused solution. \\
$y_i^\dagger, \xi_i^v, \sigma_i^2$ & Ground-truth canonical semantic target, observation noise, and noise variance used in the variance analysis. \\
\bottomrule
\end{tabular}
\end{table*}

\clearpage

\makeatletter
\setlength{\@dblfptop}{0pt plus 1fil}
\setlength{\@dblfpbot}{0pt plus 1fil}
\makeatother

\clearpage
\begin{figure*}[p]
\centering
\includegraphics[width=\textwidth,height=0.9\textheight,keepaspectratio]{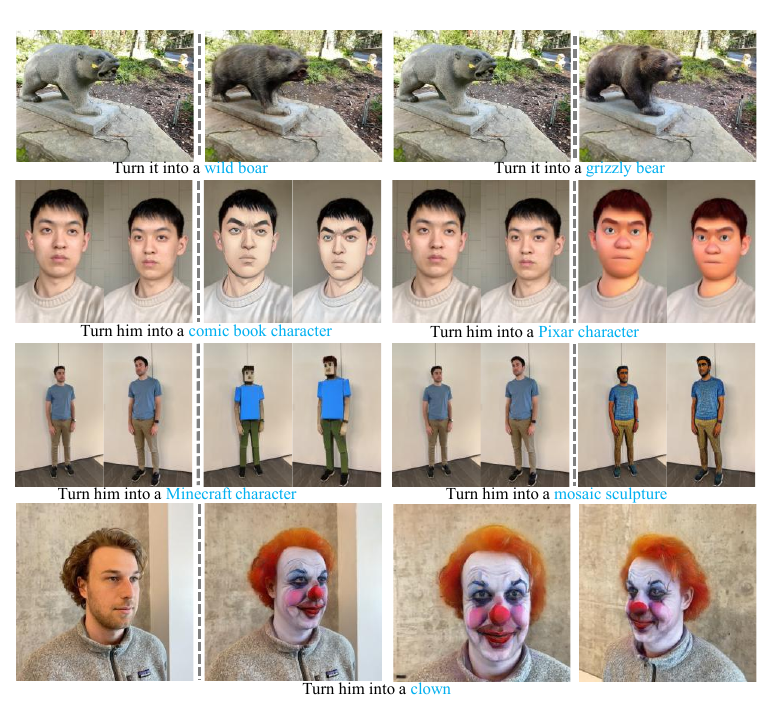}
\caption{Additional qualitative results of TransSplat (Part I).}
\Description{Representative editing results of TransSplat on pretrained 3D Gaussian Splatting scenes under text-guided instructions.}
\label{fig:more_qualitative_appendix_1}
\end{figure*}
\clearpage

\begin{figure*}[p]
\centering
\includegraphics[width=\textwidth,height=0.9\textheight,keepaspectratio]{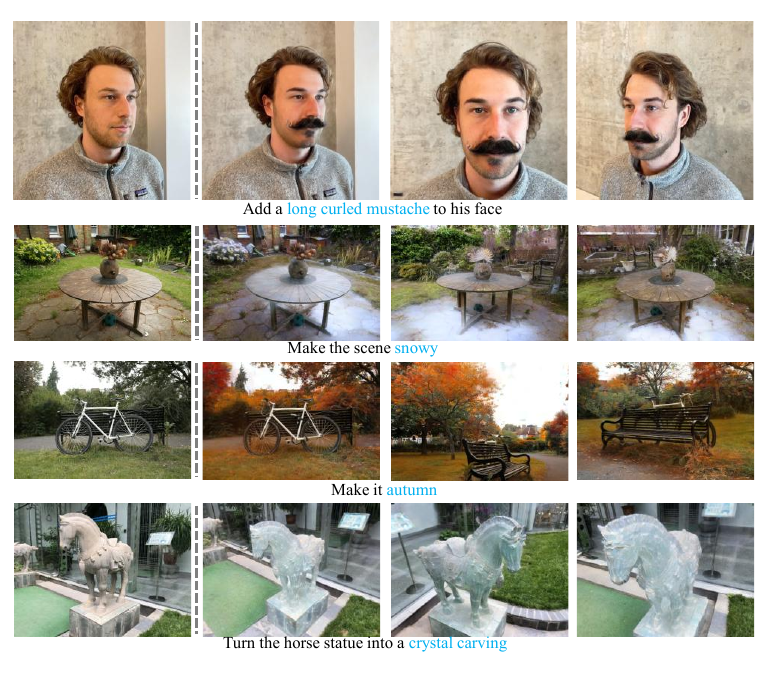}
\caption{Additional qualitative results of TransSplat (Part II).}
\Description{Additional representative editing results of TransSplat on pretrained 3D Gaussian Splatting scenes under text-guided instructions.}
\label{fig:more_qualitative_appendix_2}
\end{figure*}
\clearpage

\end{document}